\documentclass[11pt]{article}
\usepackage[letterpaper,margin=1in]{geometry}
\usepackage[T1]{fontenc}
\usepackage[utf8]{inputenc}
\usepackage{times}
\usepackage{microtype}
\usepackage{amsmath,amssymb,amsthm}
\usepackage{graphicx}
\usepackage{booktabs}
\usepackage{caption}
\usepackage{array}
\usepackage{tabularx}
\usepackage{algorithm}
\usepackage{algpseudocode}
\usepackage[numbers,sort&compress]{natbib}
\usepackage{url}
\usepackage{pgfplots}
\pgfplotsset{compat=1.16}
\usepgfplotslibrary{groupplots}
\usepackage[hidelinks]{hyperref}

\newcommand{\papertitle}{Multimodal Abstention That Preserves Acquired Correct Answers}

\newcommand{\method}{ReFrain}
\newcommand{\methodrelaxed}{ReFrain~($\varepsilon{=}0.5\%$)}
\newcommand{\Bzero}{\text{B0}}
\newcommand{\Bone}{\text{B1}}
\newcommand{\setT}{\ensuremath{\mathsf{T}}}    
\newcommand{\setCM}{\ensuremath{\mathsf{CM}}}  
\newcommand{\setE}{\ensuremath{\mathsf{E}}}    
\newcommand{\setI}{\ensuremath{\mathsf{I}}}    
\newcommand{\cand}{\mathcal{T}}                
\newcommand{\AccU}{\mathrm{Acc}_{U}}
\newcommand{\Retain}{\mathrm{Retain}}
\newcommand{\pp}{percentage points}

\theoremstyle{definition}
\newtheorem{property}{Property}
\algrenewcommand\algorithmicrequire{\textbf{Input:}}
\algrenewcommand\algorithmicensure{\textbf{Output:}}
\newcolumntype{L}[1]{>{\raggedright\arraybackslash}p{#1}}
\newcolumntype{Y}{>{\raggedright\arraybackslash}X}

\title{\papertitle}
\author{
  Jinyuan Zhang \\
  \texttt{202621116012480@stu.hubu.edu.cn}
}
\date{}

\begin{document}
\maketitle

\begin{abstract}
Adding abstention to a task-adapted multimodal model can discard correct answers, including capabilities acquired during adaptation. We formulate abstention repair as maximizing correct abstention on unanswerable questions under a budget on newly lost correct answers. We propose ReFrain (Regression-Free Abstention), which combines a linear probe on first-token logits, threshold calibration on held-out image clusters, and a deterministic veto that replaces rejected outputs with the displayed ``none of the above'' option while preserving all other responses. ``Regression-free'' denotes zero loss on the calibration set at the zero-budget target. All abstention methods in our main comparison start from the same answering-SFT checkpoint; we separately track lost correct answers, repaired errors, and retained SFT gains. On Qwen2.5-VL-3B, ReFrain at zero budget retains all 57 SFT-gained answers with no losses and improves correct abstention by 7.30 percentage points. At a 0.5\% calibration budget, it reaches 74.05\% correct abstention with 8 losses and retains 55 gains; CLIP-UP EmbLoRA reaches 64.65\% with 62 losses, 41 repairs, and 40 retained gains. The same probe at its default threshold loses 298 correct answers. With models and thresholds frozen, confirmation on 228 independent question clusters yields 66.05\% correct abstention and no losses at the 0.5\% point, exceeding EmbLoRA by 32.72 percentage points. Ordinary refusal SFT achieves higher correct abstention at greater regression cost. Code and reproduction materials: \url{https://github.com/ModeLiberaL/ReFrain}.

\end{abstract}

\section{Introduction}
\label{sec:intro}

After a multimodal model has been adapted to its task, it still needs the ability to abstain: to stop guessing on inputs where the evidence is insufficient or the question is invalid.
This order is close to practice: the model first learns to do the task, and its behavior when it cannot answer reliably is constrained afterwards.
The order also determines how the cost of the abstention stage should be counted.
Adaptation has already made the model answer a set of questions correctly, some of which adaptation has only just corrected; if the abstention stage turns these answers back into errors, overall accuracy may change very little while the specific capabilities that users can rely on have changed.
We call adding abstention to an already-adapted answer model \emph{abstention repair}, and we measure its cost against the records that this model answered correctly.

MM-UPD turns part of the answering boundary into a measurable task: the correct answer is absent from the candidate set (absent answer detection, AAD), the candidate set is incompatible with the question (incompatible answer set detection, IASD), or the image is incompatible with the question (incompatible visual question detection, IVQD)~\citep{miyai2025upd}.
In these cases, stronger domain knowledge does not necessarily make the current input solvable.
For example, knowing the general properties of an object cannot replace the corresponding evidence that is missing from the current image, and knowing the answer to the open-ended question does not make a candidate set without the correct answer yield a correct option.
Joint evaluation on Standard and UPD therefore requires both valid answering and correct abstention.

Existing approaches differ in where the abstention decision is made: continuing to train the model on abstention examples~\citep{jamil2026madmed}, injecting a visual--textual matching signal into the input~\citep{vardi2026clipup}, perturbing the input several times and deciding by the stability of the outputs~\citep{zhang2024vluncertainty}, or training a classifier that abstains directly at the output~\citep{zhao2024lvlmlp}.
TAPS separates accuracy training from post-hoc selection and compares this approach with abstention training at matched coverage~\citep{zhang2026taps}.
We study the subsequent upgrade of an already-adapted answerer: in our main comparison, all abstention schemes start from the same checkpoint obtained by supervised fine-tuning (SFT) on answering, the \emph{answer-SFT checkpoint}, and we record which answers each scheme changes relative to that checkpoint.
There are two reasons.
First, net accuracy lets losses and corrections cancel: at our common starting point, the Standard accuracy of CLIP-UP EmbLoRA is only 0.37 \pp{} below the starting point, and this difference contains 62 lost original correct answers and 41 corrections.
Second, only with the same starting point can the answers changed by each scheme be compared record by record, including the answers that answer SFT has just corrected.

Accordingly, we state abstention repair as a constrained problem: maximize the correct-abstention rate on unanswerable questions, subject to the original-correct loss not exceeding a budget~$\varepsilon$.
The original-correct loss is the negative flip rate of the model-update literature~\citep{yan2021pct,xie2021regressionbugs}.
The constrained form follows the existing formulation of regression-free model updates, in which the negative flip rate is constrained to be zero or at most a given constant~\citep{xie2021regressionbugs}.
The accompanying shared-checkpoint protocol specifies that all schemes use the same checkpoint, fit only on the training set, select operating points only on the calibration set, and neither tune thresholds nor select checkpoints on the evaluation set; besides net metrics, losses, corrections, and the retention of SFT-gained records are reported record by record.

Under this formulation we propose \method{} (Regression-Free Abstention).
It has three steps: a linear probe on the full-vocabulary logits of the first token gives an answerability score; on held-out image clusters, the largest threshold that meets the budget on the original-correct loss is selected; and when the score is below the threshold, the output is replaced with the ``none of the above'' (NOTA) option actually displayed in the question, while all other outputs stay unchanged.
The scorer is the probe of LVLM-LP.
Replacing the output directly on a classifier's decision also has precedents: LVLM-LP generates a template answer instead when its probe judges the input unanswerable, and the appendix of CLIP-UP gives a baseline that outputs a fixed refusal string after classification~\citep{zhao2024lvlmlp,vardi2026clipup}.
For post-hoc abstention baselines that threshold a confidence score on a frozen model, the threshold has been selected by an aggregate metric on a validation set~\citep{hasanov2026drno}.
\method{} calibrates the threshold against the original-correct loss of the adapted checkpoint; our controls compare this choice with the classifier's default threshold while holding the score and executor fixed.
Executing by veto only, without rewriting, gives one-sidedness: an original correct answer can be lost only by being vetoed and is never changed to another substantive option; the correct-abstention rate on unanswerable questions is not below that of the starting point; and both the loss and the correct-abstention rate are monotone in the threshold.
Calibration therefore needs only labeled answerable records, and changing the budget requires no retraining.
``Regression-free'' in the name refers to the calibration target at $\varepsilon=0$: no original correct answer is lost on the calibration set.
This is an empirical constraint on a finite calibration set.
For a new image cluster that is exchangeable with the $K$ calibration clusters, a standard order-statistic argument bounds the probability of any loss at this target by $1/(K+1)$ (Property~\ref{prop:newcluster}); the realized loss on the evaluation set is reported separately.

On the shared Qwen2.5-VL-3B~\citep{qwen25vl} answer-SFT checkpoint \Bone, \method{} loses no original correct answer on the 5,626 Standard records of the evaluation set, retains all 57 SFT-gained records, and raises the correct-abstention rate from 54.63\% to 61.94\%.
\methodrelaxed{} loses 8, with a correct-abstention rate of 74.05\%.
The same probe, abstaining at its default probability threshold, has a correct-abstention rate of 92.16\% and loses 298 original correct answers.
From the same starting point, CLIP-UP EmbLoRA loses 62 and corrects 41, with a correct-abstention rate of 64.65\%; plain abstention SFT has a correct-abstention rate of 87.47\%, loses 140 and corrects 33, and does not pass the joint criterion (Section~\ref{sec:evalprotocol}) on the evaluation set.
With calibration and the executor fixed and only the scorer replaced, the two logits probes have correct-abstention rates of 74.05\% and 75.40\% at the 0.5\% budget, and the other seven scorer configurations range from 56.59\% to 67.20\%.
On two original base models without SFT, the zero-budget point of the same pipeline loses 2 and 4 original correct answers on the evaluation set, with correct-abstention rates 2.04 and 2.85 \pp{} above EmbLoRA.

After the models and thresholds were frozen, we ran a single independent confirmation on AI2D images~\citep{kembhavi2016ai2d} that were not used in the development of this project.
Neither operating point lost an original correct answer on the 912 Standard records, and the correct-abstention rate rose by 20.10 and 44.36 \pp, respectively.
The trained branches both lose and correct answers on the confirmation set: EmbLoRA loses 7 and corrects 13; plain abstention SFT loses 4 and corrects 5, passes the joint criterion, and has a correct-abstention rate 4.90 \pp{} above \methodrelaxed.
\method{} does not correct the errors of the answer model, which is the price of one-sidedness: on the confirmation set, the point estimates of the net Standard change of the two trained branches are positive, with intervals that contain zero.

This paper makes three contributions.
First, problem and protocol: we state abstention repair after answer adaptation as a problem constrained by the original-correct loss, give a shared-checkpoint protocol and a taxonomy of schemes by where the abstention decision is made, and report the retention of SFT-gained records as a separate readout.
Second, method: \method{} and its one-sidedness properties, which reduce output-level abstention repair to a frontier parameterized by the threshold, on which the budget selects a point.
Third, evidence: on the same checkpoint we compare \method, two parameter-update schemes, and the configurations obtained by replacing the scorer; we repeat the same pipeline on two original base models; and we run a single confirmation on independent images after freezing.
The main numbers are reported as direct paired differences within the same starting point; results across base models measure how the same pipeline performs when it is fitted separately on each.

\section{Related Work}
\label{sec:related}

\paragraph{Unanswerable questions and selective prediction.}
UPD concerns inputs that are unsolvable by construction, and limits blanket abstention by evaluating each Standard question together with its unsolvable counterpart~\citep{miyai2025upd}.
Reliable VQA studies the trade-off between answer coverage and the risk of error, and ReCoVERR further examines how to reduce unnecessary abstention~\citep{whitehead2022reliable,recoverr2024}.
Together they show that the practical value of abstention depends both on stopping wrong outputs and on keeping valid answers.

\paragraph{Answer training followed by selection.}
TAPS combines accuracy-oriented post-training with a frozen answerer and an external confidence estimator~\citep{zhang2026taps}.
Its main implementation uses GRPO for accuracy training and a hidden-state correctness probe for selection; its matched-coverage analysis separates improvements in answering from improvements in rejection.
Our main experiment instead compares subsequent abstention schemes from one shared answer-SFT checkpoint, calibrates veto thresholds against its original-correct loss, and tracks the retention of records newly corrected by SFT.
This protocol measures the cost of adding abstention after task adaptation, complementing TAPS's comparison of alternative training routes.

\paragraph{Answerability and uncertainty signals.}
LVLM-LP trains probes on the information in token distributions~\citep{zhao2024lvlmlp}; VL-Uncertainty estimates the instability of outputs under visual and textual perturbations~\citep{zhang2024vluncertainty}; CLIP-UP provides a visual--textual matching signal through structured CLIP correlation vectors~\citep{vardi2026clipup}.
These methods represent different sources of information and are the direct points of comparison for understanding the differences among the operating points studied here.

\paragraph{Correctness probes and confidence fusion.}
LRP compares hidden-state and attention probes in OCR question answering~\citep{yao2025lrp}; BICR trains a correctness scorer on the difference between the states under the real image and under a black image~\citep{khanmohammadi2026bicr}; ACR fuses the maximum softmax probability (MSP) with a correctness prediction through a learnable gate~\citep{tran2026acr}.
These works provide different scoring mechanisms for a frozen answerer.
We plug multiple-choice adaptations of BICR and ACR into the calibrate and veto steps of \method{} as scorers and compare them on the same \Bone{} (Sections~\ref{sec:exp-acr}--\ref{sec:exp-bicr}).

\paragraph{Trained abstention and the evidence boundary.}
CARA studies the detection of insufficient context~\citep{liu2024cara}; MAD-MED trains with weighted abstention examples~\citep{jamil2026madmed}; SAVOR adds calibration and abstention to a reinforcement-learning reward~\citep{ding2026savor}; Variational VQA improves selective prediction through the variance of a parameter posterior~\citep{wieczorek2026varvqa}.
They cover different stages---evidence recognition, generation policy, and uncertainty-aware training---and give a broader background for applications of abstention.

\paragraph{From signal to abstention action.}
CLIP-UP injects the correlation signal into the multimodal model, in an Emb and an EmbLoRA form; its appendix also gives a classify-then-abstain baseline: a logistic regression on the correlation vector decides whether the input is answerable, a fixed refusal string is output when the input is classified as unanswerable, and the model answers otherwise.
The paper reports that this classifier labels only 46.27\% of Circular Standard questions as answerable~\citep{vardi2026clipup}.
LVLM-LP also describes probe-guided decoding: when the probe judges the input unanswerable, a template answer is generated instead~\citep{zhao2024lvlmlp}.
Dr.~No steers sparse-autoencoder features on a frozen model to induce abstention, and compares on MM-UPD a set of post-hoc abstention baselines that threshold a confidence score, with thresholds selected on a validation split to maximize mean Dual accuracy~\citep{hasanov2026drno}.
The classify-then-abstain baseline, probe-guided decoding, and the thresholded post-hoc baselines belong, like \method, to output-level override (Section~\ref{sec:taxonomy}): the answer model is frozen, and a scalar score decides whether the output is replaced.
These methods and TAPS share the external-selection architecture with \method{}.
Our experiments compare scorers and threshold rules within this architecture, using the original-correct loss of the frozen answer model to select the operating point.
Within the output-level override family we fix the executor and replace the scorer and the threshold source separately; schemes in the other families keep their own training and execution and are compared as wholes.
The two kinds of comparison answer questions about information quality and about end-to-end effect, respectively.

\paragraph{Constrained threshold selection.}
Minimizing one type of error subject to an upper bound on the other is the setting of Neyman--Pearson classification~\citep{tong2018np}.
This setting, and finite-sample methods such as conformal risk control~\citep{angelopoulos2024crc}, have been used to calibrate abstention thresholds.
For language models, what is constrained is the rate at which wrong answers are not intercepted~\citep{nie2024facttest}, or the rates at which unanswerable questions are answered and answerable ones are answered wrongly~\citep{wagner2026twoaxes}; for vision--language models, what is constrained is the error or hallucination risk on the answered part~\citep{wang2025lec,xu2026bcea}.
These works constrain erroneous outputs, and wrongly abstaining on correct answers is the side that is minimized or only reported.
Our constraint runs in the opposite direction: what is constrained is the fraction of records answered correctly by the frozen answer model that are vetoed, and what is maximized is the correct-abstention rate on unanswerable inputs.
Our calibration rule is an empirical constraint on a finite calibration set.
At a nonzero budget it gives no population-level guarantee; at the zero budget, the order-statistic arguments of these two settings bound the probability of a loss on a new exchangeable image cluster (Property~\ref{prop:newcluster}).
The realized loss on the evaluation set is reported separately, with paired intervals.

\paragraph{Regression-free updates.}
The model-update literature calls a sample that the old model answers correctly and the new model answers wrongly a negative flip, measures regression by the negative flip rate, and reduces negative flips by training the new model with distillation or by using model ensembles~\citep{yan2021pct,xie2021regressionbugs}.
Among these, \citet{xie2021regressionbugs} write the regression-free model update as a constrained optimization problem, constraining the negative flip rate to be zero or at most a given constant, and then relax it into joint training with a distillation term.
Another line keeps the predictions of the old model at inference time and chooses between, or fuses, the old and new predictions sample by sample~\citep{trauble2021backward,lai2023gatedfusion}.
For language models, negative flips are also used to measure the regression caused by model version updates and by preference alignment~\citep{echterhoff2024muscle,zhu2024flipguard}.
We adopt the negative flip rate as the measure together with the constrained form, and change the object to the addition of abstention: the reference model is a checkpoint that has completed answer adaptation, and the objective is the correct-abstention rate on unanswerable inputs.
\method{} neither trains a new answer model nor chooses between two answer models; there is only one answer model, the only alternative output is abstention, and the veto threshold is calibrated against the original-correct loss on held-out image clusters.
In the post-training setting, we additionally condition on the records that the previous stage newly made correct, to characterize how a capability is retained after it is acquired.

\section{Problem, Protocol, and Method}
\label{sec:method}

\subsection{Problem: Abstention Repair after Answer SFT}
\label{sec:problem}

\paragraph{Records and correctness.}
A record $x$ consists of an image, a question, and a set of candidate options, among which a ``none of the above'' (NOTA) option is displayed.
$S$ denotes the set of Standard records, whose correct answer is a substantive option and on which a wrongful abstention counts as an error.
$U$ denotes the set of UPD records, on which an abstention that conforms to the task definition counts as correct; selecting the displayed NOTA option is such an abstention.
$c_g(i)\in\{0,1\}$ indicates whether system $g$ is correct on record $i$.

\paragraph{Answer model.}
Let \Bzero{} be the original multimodal answer model and \Bone{} the model after a fixed answer SFT.
Our main setting takes $f=\Bone$ and freezes its parameters.
This corresponds to a common order of use: the model is first adapted to the task and only afterwards given the ability to abstain, and the cost of the abstention stage is counted against the records that this model answered correctly.
Section~\ref{sec:exp-base} additionally takes $f$ to be each of two original base models, to examine the same pipeline from starting points without SFT.

\paragraph{Abstention repair schemes.}
An abstention repair scheme takes $f$, a training set \setT, and a calibration set \setCM, and outputs a repaired system $g$.
\setT{} is used for fitting or training and \setCM{} for selecting an operating point or a model; the evaluation set takes part in no selection.

\paragraph{Loss and correction.}
The original-correct loss and the correction rate of $g$ relative to $f$ are
\begin{equation}
L(g;f)=\frac{1}{|S|}\sum_{i\in S}c_f(i)\,[1-c_g(i)],\qquad
F(g;f)=\frac{1}{|S|}\sum_{i\in S}[1-c_f(i)]\,c_g(i).
\label{eq:loss-correction}
\end{equation}
The net change in accuracy satisfies $\Delta\mathrm{Acc}=F-L$.
$L$ is the fraction of records that $f$ answers correctly and $g$ answers wrongly, that is, the negative flip rate of the model-update literature~\citep{yan2021pct,xie2021regressionbugs}; $F$ is the fraction in the opposite direction.

\paragraph{Constrained problem.}
Given a budget $\varepsilon\ge 0$, abstention repair solves
\begin{equation}
\max_{g}\ \AccU(g)\quad\text{s.t.}\quad L(g;f)\le\varepsilon,
\label{eq:problem}
\end{equation}
where $\AccU$ is the correct-abstention rate on $U$.
$\varepsilon=0$ requires that no original correct answer be lost and is the preset primary operating point; $\varepsilon=0.5\%$ is a preset additional operating point.
Writing the regression-free model update as a constrained optimization problem, with the negative flip rate constrained to be zero or at most a given constant, has a precedent~\citep{xie2021regressionbugs}; we adopt this form, restrict the update to adding abstention to an already-adapted answer model, and take the objective to be the correct-abstention rate on unanswerable records.

The constraint is placed on $L$ rather than on net accuracy because the net value lets losses and corrections cancel.
One trained branch in Section~\ref{sec:exp-b1} is only 0.37 \pp{} below $f$ on Standard, and this difference contains 62 lost original correct answers and 41 corrections.
For a model that is already in use, what users encounter are these record-level regressions.

\paragraph{SFT-gained records.}
Define $R=\{i\in S: c_{\Bzero}(i)=0,\ c_{\Bone}(i)=1\}$, the records that answer SFT has just corrected.
The gain retention rate is
\begin{equation}
\Retain(g;R)=\frac{\sum_{i\in R}c_g(i)}{|R|}.
\label{eq:retain}
\end{equation}
It answers whether the abstention stage concentrates its losses on the capability gained in the previous stage.
$L$, $F$, and $\Retain$ have different denominators and meanings and are listed separately in the results.

\subsection{Shared-Checkpoint Protocol}
\label{sec:protocol}

Answer SFT uses the 677 Standard records in \setT, trains for three epochs, and saves a single final checkpoint, \Bone.
All abstention repair schemes start from this one \Bone.

\setT{} has 1,354 records.
The calibration set \setCM{} has 4,567 records in 183 image clusters, of which 2,406 are Standard.
The evaluation set \setE{} has 10,678 records in 416 image clusters, of which 5,626 are Standard and 5,052 are UPD.
The image-cluster split isolates training, calibration, and evaluation, and the image cluster is the resampling unit of the paired intervals.

All schemes follow the same set of rules (Table~\ref{tab:protocol}).
When the starting point is an original base model (Section~\ref{sec:exp-base}), the same protocol applies, and scorers and thresholds are fitted separately on each base model.

\begin{table}[ht]
\centering
\caption{Rules of the shared-checkpoint protocol.}
\label{tab:protocol}
\small
\begin{tabularx}{0.92\linewidth}{@{}L{4.6cm}Y@{}}
\toprule
Resource & Rule \\
\midrule
Answering starting point & The same \Bone{} checkpoint \\
Fitting or training & \setT{} only \\
Selection of operating point or model & \setCM{} only \\
Evaluation & \setE{} and the confirmation set \setI{} (Section~\ref{sec:confirmation-setup}); no threshold tuning or checkpoint selection on them \\
Parsing and decoding & The same parser, the same NOTA mapping, greedy decoding \\
Readouts & All readouts of Section~\ref{sec:evalprotocol} \\
\bottomrule
\end{tabularx}
\end{table}

Each compared scheme carries a fidelity label.
A \emph{full reproduction} uses the official code and all components.
A \emph{mechanism port} implements the mechanism of the original paper in our setting and discloses the hyperparameters that the paper does not give.
A \emph{partial implementation} lacks a component of the original method and does not enter the main tables under the name of the original method.

\subsection{A Taxonomy of Abstention Repair Schemes}
\label{sec:taxonomy}

By whether the answer model is modified and by where the abstention decision is made, existing approaches fall into six families (Table~\ref{tab:taxonomy}).

\begin{table}[ht]
\centering
\caption{Families of abstention repair schemes. ``Modifies $f$'' states whether the parameters of $f$ are changed; ``Can rewrite'' states whether a correct answer of $f$ can be changed to another substantive option.}
\label{tab:taxonomy}
\footnotesize
\begin{tabularx}{\linewidth}{@{}L{2.1cm}L{3.5cm}L{2.0cm}L{1.7cm}Y@{}}
\toprule
Family & What it does & Modifies $f$ & Can rewrite & Representative methods \\
\midrule
Output-level override & Freezes $f$; a scalar score decides whether the output is replaced with NOTA & No & No & \method{} (this paper); the LVLM-LP probe abstaining at its default threshold~\citep{zhao2024lvlmlp}, the CLIP-UP correlation-vector classifier~\citep{vardi2026clipup}, BICR~\citep{khanmohammadi2026bicr}, ACR~\citep{tran2026acr} \\
\addlinespace
Parameter update & Continues training $f$ on abstention examples & Yes & Yes & Plain abstention SFT, CLIP-UP EmbLoRA (which also injects CLIP correlation vectors)~\citep{vardi2026clipup}, MAD-MED~\citep{jamil2026madmed}, SAVOR~\citep{ding2026savor}, VL-Calibration~\citep{xiao2026vlcalibration}, Reliable-V-RL~\citep{jian2026reliablevrl} \\
\addlinespace
Input or representation injection & Adds a signal to the input embeddings or to intermediate activations & Only the added projection or steering parameters & Yes & CLIP-UP Emb~\citep{vardi2026clipup}, Dr.~No~\citep{hasanov2026drno} \\
\addlinespace
Inference-time sampling & Perturbs the input several times and decides by the consistency of the outputs & No & Depends on the execution & VL-Uncertainty~\citep{zhang2024vluncertainty} \\
\addlinespace
Learned routing & Trains the model to output an answer-or-abstain decision before answering & Yes & Yes & Rep2Act~\citep{rep2act2026} \\
\addlinespace
Prompting & Changes the instruction only & No & Yes & The additional-instruction setting of MM-UPD~\citep{miyai2025upd} \\
\bottomrule
\end{tabularx}
\end{table}

Within the output-level override family there are two layers of comparison.
The first fixes the score and changes the source of the threshold: the same score is executed at the default threshold of the original method, or calibrated by the rule of Section~\ref{sec:refrain}.
The second fixes calibration and the executor and replaces the score.
Schemes in the other families are compared as wholes, together with their own training and execution.

On the same \Bone{} we evaluate the scorers of the output-level override family (Sections~\ref{sec:exp-scorers}--\ref{sec:exp-bicr}) and two schemes of the parameter-update family (Section~\ref{sec:exp-b1}).
CLIP-UP Emb, from the input-injection family, and the inference-time sampling method are evaluated on the original base models (Section~\ref{sec:exp-base}).
Dr.~No, the learned-routing family, and the prompting family have no same-setting results yet (Section~\ref{sec:limitations}).

\subsection{ReFrain}
\label{sec:refrain}

\method{} (Regression-Free Abstention) belongs to the output-level override family and consists of three steps: score, calibrate, and veto (Algorithm~\ref{alg:refrain}).
It changes neither the parameters of $f$ nor the substantive answers of $f$.

\begin{algorithm}[ht]
\caption{\method}
\label{alg:refrain}
\begin{algorithmic}[1]
\Require frozen answer model $f$; training set \setT{} with Standard/UPD labels; the Standard records of \setCM{} and the correctness of $f$ on them; budget $\varepsilon$
\Ensure $g_{\tau_\varepsilon}$
\State \textbf{Score.} Fit a scorer $s$ on \setT; a low score indicates that abstention is more warranted.
\State \textbf{Calibrate.} $\tau_\varepsilon \gets \max\{\tau\in\cand:\ \widehat{L}_{\setCM}(g_\tau;f)\le\varepsilon\}$, where $\cand=\{-\infty,+\infty\}\cup\{s(x):x\in\setT\}$.
\State \textbf{Veto.} $g_{\tau_\varepsilon}(x)\gets n(x)$ if $s(x)<\tau_\varepsilon$; otherwise $g_{\tau_\varepsilon}(x)\gets f(x)$.
\end{algorithmic}
\end{algorithm}

\paragraph{Score.}
The default scorer is a linear probe.
It reads the full-vocabulary logits $\phi_f(x)$ at the position where $f$ begins to generate its answer and outputs an answerability score $s(x)=w^\top\phi_f(x)+b$.
The probe is trained on \setT{} with Standard as the positive class and UPD as the negative class; the regularization strength is selected by five-fold cross-validation over image clusters within \setT, with mean binary cross-entropy as the criterion.
The feature and the form of the probe follow LVLM-LP~\citep{zhao2024lvlmlp}.
$\phi_f(x)$ comes from the same forward pass that produces the answer, so scoring adds no forward passes.

\paragraph{Calibrate.}
The candidate threshold set is $\cand=\{-\infty,+\infty\}\cup\{s(x):x\in\setT\}$, including the no-veto and all-veto endpoints.
For a budget $\varepsilon$,
\begin{equation}
\tau_\varepsilon=\max\{\tau\in\cand:\ \widehat{L}_{\setCM}(g_\tau;f)\le\varepsilon\},
\label{eq:calibrate}
\end{equation}
where $\widehat{L}_{\setCM}$ is the empirical original-correct loss on \setCM, whose denominator is all 2,406 Standard records of \setCM.
$\varepsilon=0.5\%$ allows at most 12 calibration losses, and the default scorer attains 11.
The threshold is determined directly by how many original correct answers may be lost, and not by the default probability threshold of a classifier, an aggregate metric, or an error risk.
Under veto-only execution, this rule is the empirical form of Neyman--Pearson threshold selection: the constrained error is the veto of a record that $f$ answered correctly.
Selecting a threshold by the empirical error rate does not guarantee with high probability that the population error rate stays within the same level~\citep{tong2018np}; we therefore report the constraint on the calibration set and the realized loss on the evaluation set separately.
For $\varepsilon=0$, Property~\ref{prop:newcluster} states the level that does hold for a new image cluster under an exchangeability assumption.

\paragraph{Veto.}
Let $n(x)$ be the NOTA letter actually displayed in the question, and
\begin{equation}
g_\tau(x)=
\begin{cases}
n(x), & s(x)<\tau,\\
f(x), & \text{otherwise.}
\end{cases}
\label{eq:veto}
\end{equation}
The system has only a veto: it can replace the output of $f$ with NOTA, but not with another substantive option.
Records that are not vetoed keep the original output of $f$.
Replacing the output on a classifier's decision has precedents: the probe-guided decoding of LVLM-LP generates a template answer instead when the input is judged unanswerable, and the classification baseline in the appendix of CLIP-UP outputs a fixed refusal string~\citep{zhao2024lvlmlp,vardi2026clipup}.
What \method{} substitutes is the NOTA option displayed in the question.

\paragraph{The scorer is a replaceable component.}
The first step can be replaced with any scalar score, with calibration and veto unchanged.
We use this to compare, under the same calibration and executor, a probe with a fixed regularization strength ($C=1$), the maximum softmax probability, the negative entropy, the CLIP correlation-vector classifier, and the scores of BICR and ACR (Sections~\ref{sec:exp-scorers}--\ref{sec:exp-bicr}).
Of these, BICR and ACR predict whether the answer of $f$ is correct, whereas the probes and the CLIP classifier predict whether the input is answerable.

\paragraph{Name and notation.}
In what follows, \method{} without annotation denotes the default configuration: the cross-validated probe at $\varepsilon=0$.
``Regression-free'' in the name refers to this calibration target: not a single record that $f$ answered correctly is lost on the calibration set.
It is an empirical constraint on a finite calibration set.
For a new image cluster that is exchangeable with the calibration clusters, Property~\ref{prop:newcluster} bounds the probability of any loss; for data from other sources there is no guarantee, and the realized loss on the evaluation set is reported separately.
The operating point at $\varepsilon=0.5\%$ is written \methodrelaxed{} and is the relaxed version.
When the scorer is replaced, we write \method~(scorer name).
The names of the original methods, such as LVLM-LP, CLIP-UP, BICR, and ACR, are used only for results obtained with the original method's own threshold and execution.

\subsection{Properties}
\label{sec:properties}

The following two properties depend only on the veto rule $g_\tau(x)\in\{f(x),n(x)\}$ and hold for any scorer.

\begin{property}[One-sidedness]
\label{prop:onesided}
(a) For every Standard record, $c_g\le c_f$.
Hence $F(g_\tau;f)=0$ and $\Delta\mathrm{Acc}=-L$: a veto does not correct an error of $f$, nor does it change a correct answer of $f$ to another substantive option; the only cost is the original correct answers that are vetoed.
(b) For every UPD record, $c_g\ge c_f$; the correct-abstention rate is not below that of $f$ at any threshold.
(c) As $\tau$ increases, the set of vetoed records only grows, and $L(g_\tau;f)$ and $\AccU(g_\tau)$ are both non-decreasing in $\tau$.
The whole method is therefore a frontier parameterized by $\tau$, on which the budget $\varepsilon$ selects a point (Section~\ref{sec:exp-curves}).
\end{property}

Part~(a) holds because NOTA is always a wrong answer on a Standard record; part~(b) holds because the displayed NOTA option is always a correct answer on a UPD record.
Parameter-update schemes have neither (a) nor (b): they can lose and correct Standard answers at the same time, and can fall below $f$ on some UPD records.

\begin{property}[The loss is determined by one score distribution]
\label{prop:loss}
\begin{equation}
L(g_\tau;f)=\frac{1}{|S|}\sum_{i\in S}c_f(i)\,\mathbf{1}[s(x_i)<\tau].
\label{eq:loss-score}
\end{equation}
\end{property}

The loss equals the Standard accuracy of $f$ multiplied by the fraction of the Standard records answered correctly by $f$ whose score is below $\tau$.
Calibration is therefore the selection of a point on a monotone curve, and it needs only answerable records with their original task labels and the correctness of $f$ on them; unanswerable examples are used only to train the scorer.
Changing the budget only requires selecting the threshold again, not retraining.
At $\varepsilon=0$, $\tau_0$ is the largest candidate threshold that does not exceed the lowest score among the \setCM{} Standard records that $f$ answers correctly.
When $f=\Bone$, the gain retention rate is the restriction of the same quantity to $R$:
\begin{equation}
\Retain(g_\tau;R)=1-\frac{|\{i\in R:\ s(x_i)<\tau\}|}{|R|}.
\label{eq:retain-score}
\end{equation}

The third property concerns new data and, unlike the first two, needs a sampling assumption.
For an image cluster $k$, let $m_k$ be the lowest score among its Standard records that $f$ answers correctly, with $m_k=+\infty$ if there is none.
By Property~\ref{prop:loss}, $g_\tau$ loses an original correct answer in cluster $k$ exactly when $m_k<\tau$, and the zero-budget rule gives $\tau_0\le\min_{k\le K}m_k$ over the $K$ image clusters of \setCM.

\begin{property}[Zero-budget loss on a new image cluster]
\label{prop:newcluster}
Suppose that $f$ and the scorer $s$ are fixed without reference to the image clusters of \setCM{} or to the new image cluster, and that, conditional on the training set and on everything else that determines $f$ and $s$, the $K$ image clusters of \setCM{} and the new image cluster are exchangeable.
(a) The probability that $g_{\tau_0}$ loses at least one original correct answer in the new cluster is at most $1/(K+1)$.
(b) If, under the same conditioning, the clusters are moreover i.i.d., then for every $\alpha\in(0,1)$, with probability at least $1-(1-\alpha)^K$ over the calibration clusters, the probability of such a loss in a new cluster is at most $\alpha$.
\end{property}

Both parts follow from a standard order-statistic argument and are stated for completeness, not as a new result.
Part~(a) is the case of conformal risk control in which the target level is $1/(K+1)$, so that the rule requires zero empirical risk~\citep{angelopoulos2024crc}; part~(b) is the bound for the extreme order statistic in the Neyman--Pearson umbrella algorithm~\citep{tong2018np}.
Proofs are given in Appendix~\ref{app:proofs}.
With $K=183$, part~(a) gives $1/184\approx 0.54\%$, and part~(b) gives $\alpha\approx 1.62\%$ at the 95\% level.
The probability in part~(a) is over the joint draw of the calibration clusters and the new cluster and is not conditional on the realized calibration set; part~(b) is the conditional statement.

The scope of Property~\ref{prop:newcluster} is narrow.
It is stated for image clusters and not for records; it bounds the probability for a single new cluster and is not simultaneous over the clusters of an evaluation set; it covers only $\varepsilon=0$ and says nothing about \methodrelaxed; and it does not apply to data from another source, such as the confirmation set.
In the protocol of Section~\ref{sec:protocol}, $f$ and the default scorer are fitted on \setT{} only.
For \setE, exchangeability with the image clusters of \setCM{} is an assumption about how the data were partitioned; it holds by design when image clusters are assigned to \setCM{} and \setE{} uniformly at random.
Under it, the expected number of the 416 image clusters of \setE{} in which an original correct answer is lost is at most $416/184\approx 2.26$; the observed number is 0 for $f=\Bone$ (Section~\ref{sec:exp-b1}).
Both sets were also used while the method was developed, so the first condition is not verified for the design choices; the observed number is descriptive, a consistency check and not a test of the bound.

\paragraph{Cost.}
One-sidedness means that \method{} gives up correction, and its Standard accuracy cannot be higher than that of $f$.
Schemes that can correct the errors of $f$ can be higher than $f$: on the confirmation set, the net Standard point estimates of the two trained branches are both above $f$, with intervals that contain zero (Section~\ref{sec:exp-confirm}).
The gain in the correct-abstention rate equals the fraction of UPD records that $f$ answered wrongly and whose score is below $\tau$, so the ceiling on abstention coverage is set by how well the scorer separates the two kinds of records.

\subsection{Evaluation Protocol}
\label{sec:evalprotocol}

\paragraph{Readouts.}
We report Standard accuracy, the UPD correct-abstention rate, Circular Dual, the number of original correct answers lost, the number of corrections, and the gain retention rate.
Circular aggregates over the permutations of the candidates; Dual requires that the Standard side and the UPD side of the same question both satisfy their conditions.

\paragraph{Joint criterion.}
Relative to the common starting point $f$, the lower bound of the paired 95\% interval of $\Delta$Standard is above $-1$ percentage point, and at the same time the lower bound for $\Delta$UPD is above zero.
The intervals use 2,000 paired bootstrap resamples over image clusters~\citep{efron1979bootstrap}.
An advantage of one scheme over another is tested separately with direct paired differences.
Passing the criterion relative to $f$ and being better than another scheme are reported separately.

\paragraph{Native view.}
Each scheme is reported at its own default operating point, with all readouts and the joint criterion.

\subsection{Independent Confirmation of the Frozen Setting}
\label{sec:confirmation-setup}

The confirmation set \setI{} comes from the non-test images of the complete official AI2D image collection~\citep{kembhavi2016ai2d}.\footnote{\url{https://allenai.org/data/diagrams}}
Its construction excluded the images of the project's earlier MM-UPD and CLIP-UP data, the complete official AI2D test images, and the images reserved by a parallel study, and also excluded exact and near-duplicate images.
The original questions keep their official answers; AAD removes the correct option, IASD replaces the option set, and IVQD replaces the image with a mismatched foreign image.
After a review of text and image semantics by fixed rules, 228 of 300 candidate clusters were kept: 96 AAD, 89 IASD, and 43 IVQD.
Standard records rotate over all four substantive options, AAD records over three, and the other UPD records over four; NOTA is fixed in the last position.
The final set has 271 images, 912 Standard records, and 816 UPD records.
Each IVQD foreign image is used in only one cluster and is resampled together with all records of the original question.

\Bone, the EmbLoRA and plain abstention SFT checkpoints, the probe coefficients, and the two \setCM{} thresholds were frozen before inference on the confirmation set; the thresholds are $-10.226138449867452$ and $-6.183404341319426$.
The confirmation set takes no part in training or calibration, and neither the selection nor the semantic review read the predictions of any branch under evaluation.
\Bzero{} only generates Standard answers, to identify the SFT-gained records.
All comparisons keep the original parser, greedy decoding, and the joint criterion relative to \Bone; the two operating points and the four paired comparisons between methods were fixed before inference.
The results are reported separately from \setE, with both the UPD micro-average and the task macro-average.
Details of the construction are given in Appendix~\ref{app:confirm}.

\section{Results}
\label{sec:experiments}

Unless stated otherwise, the numbers in this section are measured on \setE, differences between rates are given in percentage points (pp), and intervals are paired 95\% intervals over image clusters (Section~\ref{sec:evalprotocol}).
In the tables, UPD is the correct-abstention rate on UPD records, ``Lost'' and ``Corrected'' count Standard records relative to the starting point $f$, and ``Retained'' counts SFT-gained records.

\subsection{Two Original Base Models under the Same Protocol}
\label{sec:exp-base}

Table~\ref{tab:base} takes $f$ to be each of two original base models without answer SFT, Qwen2.5-VL-3B and InternVL3-8B~\citep{zhu2025internvl3}.
The zero budget is a constraint on \setCM.
On \setE, the zero-budget points on the two base models lose 2 and 4 original correct Standard answers, and their correct-abstention rates are 2.04 and 2.85 pp above EmbLoRA.
The probe and the threshold are fitted separately on each base model.
These losses are consistent with the scope of Property~\ref{prop:newcluster}, which bounds the probability of a loss in a new image cluster and does not state that the loss on the evaluation set is zero.

\begin{table}[ht]
\centering
\caption{Two original base models without answer SFT, on \setE.
Each scheme is fitted and calibrated separately on each base model, and the joint criterion is relative to the base model without an abstention stage.
\method{} takes $f$ to be the base model.
VL-Uncertainty is reported with its native execution and with its score under the common executor.}
\label{tab:base}
\small
\begin{tabular}{@{}llrrl@{}}
\toprule
Original base model & Scheme & Standard (\%) & UPD (\%) & Joint criterion \\
\midrule
Qwen2.5-VL-3B & No abstention stage & 86.72 & 56.43 & reference \\
 & \method & 86.69 & 67.85 & passed \\
 & CLIP-UP EmbLoRA & 86.76 & 65.82 & passed \\
 & CLIP-UP Emb & 86.21 & 68.01 & not passed \\
 & VL-Uncertainty (native execution) & 80.16 & 75.36 & not passed \\
 & VL-Uncertainty score, common executor & 86.70 & 56.63 & passed \\
\midrule
InternVL3-8B & No abstention stage & 92.18 & 74.33 & reference \\
 & \method & 92.11 & 80.84 & passed \\
 & CLIP-UP EmbLoRA & 92.09 & 77.99 & passed \\
 & CLIP-UP Emb & 80.48 & 53.35 & not passed \\
\bottomrule
\end{tabular}

\end{table}

CLIP-UP~\citep{vardi2026clipup} behaves differently across base models and forms of attachment.
EmbLoRA passes the joint criterion on both base models; Emb degrades markedly on InternVL3-8B under the common prompt.
This observation corresponds to the present combination of input, injection, and training settings, and we do not read it as a statement about the original method in its own setting (Appendix~\ref{app:fidelity}).
The native execution of VL-Uncertainty~\citep{zhang2024vluncertainty} and its score under the common executor give clearly different operating points and are therefore reported separately.

\subsection{A Common Starting Point after Answer SFT}
\label{sec:exp-b1}

Answer SFT corrects 57 answers that \Bzero{} got wrong and loses 29 that \Bzero{} got right, a net gain of 28 correct answers.
The 57 SFT-gained records lie in 28 image clusters.
Table~\ref{tab:b1} gives the results of the abstention branches that start from \Bone.

\begin{table}[ht]
\centering
\caption{Abstention branches from the common starting point \Bone, on \setE.
Lost and corrected answers are counted relative to \Bone{} among the 5,626 Standard records; retained records are out of the 57 SFT-gained records.}
\label{tab:b1}
\small
\begin{tabular}{@{}lrrrrr@{}}
\toprule
Branch & Standard (\%) & UPD (\%) & Circular Dual (\%) & Lost / corrected & Retained \\
\midrule
\Bzero & 86.72 & 56.43 & 40.26 & --- & --- \\
\Bone & 87.22 & 54.63 & 38.86 & 0 / 0 & 57 / 57 \\
\method & 87.22 & 61.94 & 46.22 & 0 / 0 & 57 / 57 \\
\methodrelaxed & 87.08 & 74.05 & 55.73 & 8 / 0 & 55 / 57 \\
CLIP-UP EmbLoRA & 86.85 & 64.65 & 47.82 & 62 / 41 & 40 / 57 \\
Plain abstention SFT & 85.32 & 87.47 & 66.41 & 140 / 33 & 23 / 57 \\
\bottomrule
\end{tabular}

\end{table}

Both \method{} operating points and EmbLoRA pass the joint criterion relative to \Bone.
\method{} keeps every original correct answer of \Bone{} and raises the correct-abstention rate by 7.30 pp, with a 95\% interval of [6.21, 8.37].
Its correct-abstention rate is 2.71 pp below that of EmbLoRA, with a direct paired interval of [$-$4.07, $-$1.32], which shows a trade-off between stricter retention and abstention coverage.

The evaluation-set loss of \methodrelaxed{} is 8 of 5,626 Standard records, of which 2 are SFT-gained records and 6 are other records that \Bone{} answered correctly.
Relative to EmbLoRA, its correct-abstention rate is 9.40 pp higher, with an interval of [7.77, 11.05]; it loses 54 fewer original correct answers, a difference of 0.96 pp with an interval of [0.52, 1.47].
The point estimate of its Standard accuracy is 0.23 pp higher, with an interval of [$-$0.37, 0.87].

The gain retention rate is 96.49\% against 70.18\%, a direct difference of 26.32 pp with an interval of [7.35, 49.06].
EmbLoRA loses 17 SFT-gained records and 45 other correct records, and corrects 41 records that \Bone{} answered wrongly.
The corrections are a real gain of this branch; the two kinds of transition together make up net accuracy.

Plain abstention SFT starts from the same \Bone, uses the same 1,354 training records and three epochs of training, and takes the LoRA configuration and the learning rate of EmbLoRA.
Its target is the correct option or the NOTA letter in the common multiple-choice format.
This branch has a higher correct-abstention rate and a higher Circular Dual than the other branches in Table~\ref{tab:b1}; it loses 34 SFT-gained records and 106 other answers that \Bone{} got right, and corrects 33 old errors.
Its Standard accuracy falls by 1.90 pp relative to \Bone, with a 95\% interval of [$-$2.84, $-$1.01], so it does not pass the accuracy part of the joint criterion.

\begin{figure}[ht]
\centering
\definecolor{figink}{HTML}{184F95}
\definecolor{figlost}{HTML}{184F95}
\definecolor{figcorrected}{HTML}{EB6834}
\begin{tikzpicture}
\begin{groupplot}[
  group style={group size=3 by 1, horizontal sep=0.8cm, yticklabels at=edge left},
  scale only axis, width=3.4cm, height=4.4cm,
  xbar, xmin=0, y dir=reverse, ymin=0.4, ymax=6.6,
  ytick={1,...,6}, yticklabels={{\method},{\methodrelaxed},{\method~(BICR)$^{\dagger}$},{\method~(ACR-MCQ)$^{\dagger}$},{CLIP-UP EmbLoRA},{Plain abstention SFT}},
  yticklabel style={font=\footnotesize}, xticklabel style={font=\footnotesize},
  title style={font=\footnotesize, align=center, text width=3.4cm},
  axis x line*=bottom, axis y line*=left, ytick style={draw=none},
  xmajorgrids, grid style={black!12},
  nodes near coords, point meta=explicit symbolic,
  nodes near coords align={right},
  nodes near coords style={font=\scriptsize, text=black, inner sep=2pt},
  clip=false,
]
\nextgroupplot[title={(a) Correct-abstention rate (\%)}, xmax=100, xtick={0,25,50,75,100}, bar width=7pt]
\addplot[fill=figink, draw=none] coordinates {(61.94,1) [61.94] (74.05,2) [74.05] (58.23,3) [58.23] (67.20,4) [67.20] (64.65,5) [64.65] (87.47,6) [87.47]};
\draw[dashed, black!70] (axis cs:54.63,0.4) -- (axis cs:54.63,6.6);
\nextgroupplot[title={(b) Records lost and corrected}, xmax=170, xtick={0,50,100,150}, bar width=5pt,
  legend style={font=\scriptsize, draw=none, fill=none, at={(0.98,0.98)}, anchor=north east},
  legend cell align=left, legend image code/.code={\draw[#1, draw=none] (0cm,-0.08cm) rectangle (0.28cm,0.08cm);}]
\addplot[fill=figlost, draw=none, bar shift=3.5pt] coordinates {(0,1) [0] (8,2) [8] (10,3) [10] (36,4) [36] (62,5) [62] (140,6) [140]};
\addplot[fill=figcorrected, draw=none, bar shift=-3.5pt] coordinates {(0,1) [0] (0,2) [0] (0,3) [0] (0,4) [0] (41,5) [41] (33,6) [33]};
\legend{Lost, Corrected}
\nextgroupplot[title={(c) SFT-gained records retained}, xmax=66, xtick={0,19,38,57}, bar width=7pt]
\addplot[fill=figink, draw=none] coordinates {(57,1) [57] (55,2) [55] (57,3) [57] (53,4) [53] (40,5) [40] (23,6) [23]};
\end{groupplot}
\end{tikzpicture}
\caption{Results on the same \Bone, on \setE.
(a) Correct-abstention rate on UPD records; the dashed line marks \Bone{} (54.63\%).
(b) Original correct answers lost and old errors corrected, relative to \Bone.
(c) SFT-gained records retained, out of 57.
$^{\dagger}$The scorer is replaced with that of BICR or of ACR-MCQ (standardized features) and calibrated at the 0.5\% budget under the common executor (Sections~\ref{sec:exp-acr} and~\ref{sec:exp-bicr}).}
\label{fig:b1}
\end{figure}
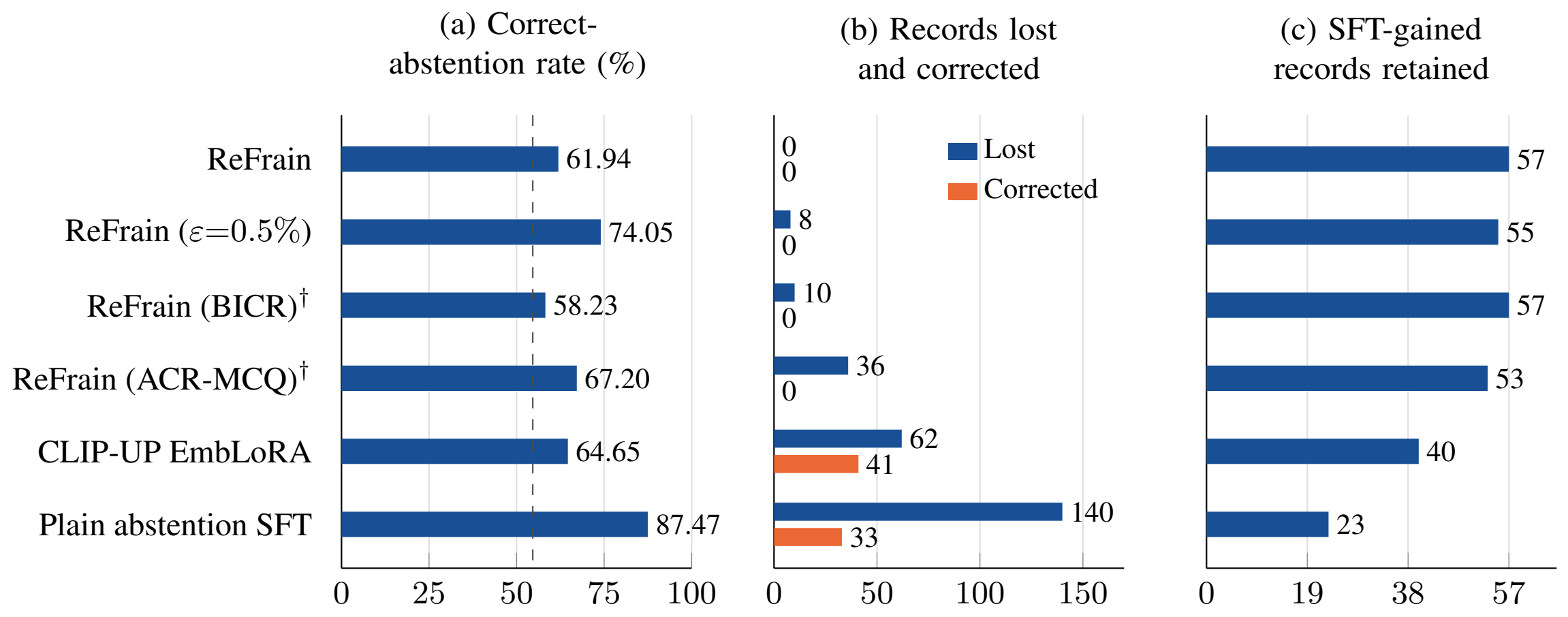

Relative to plain abstention SFT, \methodrelaxed{} is 1.76 pp higher on Standard, with an interval of [0.89, 2.68], and 13.42 pp lower in correct-abstention rate, with an interval of [$-$14.88, $-$11.90].
The two offer, respectively, stronger retention of correct behavior and higher abstention coverage.
Figure~\ref{fig:b1} shows the abstention gain, the transitions of correct answers, and the retention of SFT-gained records together.

After SFT, the correct-abstention rate of \Bone{} is 1.80 pp below that of \Bzero.
Relative to \Bzero, the final correct-abstention rates of \method, \methodrelaxed, and EmbLoRA are higher by 5.50, 17.62, and 8.21 pp.
The final gain of the abstention stage therefore exceeds the recovery of the drop in abstention caused by this SFT.

\subsection{Scorers under the Same Calibration and Executor}
\label{sec:exp-scorers}

Table~\ref{tab:scorers} replaces the scorer and keeps the calibration rule and the executor.
The ordering of the probe with fixed $C=1$ and the cross-validated probe changes with the budget.
At $\varepsilon=0$, the correct-abstention rate of the cross-validated probe is 2.06 pp higher, with an interval of [1.58, 2.58]; at $\varepsilon=0.5\%$, that of the $C=1$ probe is 1.35 pp higher, with an interval of [0.86, 1.84], and this probe loses 5 more original correct answers.
The two probes retain the same number of SFT-gained records.

\begin{table}[ht]
\centering
\caption{Scorers under the same calibration and executor on \Bone, on \setE.
MSP is the maximum softmax probability.
Every row is \method{} with the named scorer.}
\label{tab:scorers}
\small
\begin{tabular}{@{}lrrrrr@{}}
\toprule
Scorer & $\varepsilon$ (\%) & Standard (\%) & UPD (\%) & Lost & Retained \\
\midrule
Cross-validated probe (default) & 0 & 87.22 & 61.94 & 0 & 57 / 57 \\
Probe with fixed $C=1$ & 0 & 87.22 & 59.88 & 0 & 57 / 57 \\
Cross-validated probe (default) & 0.5 & 87.08 & 74.05 & 8 & 55 / 57 \\
Probe with fixed $C=1$ & 0.5 & 86.99 & 75.40 & 13 & 55 / 57 \\
MSP & 0.5 & 86.97 & 59.34 & 14 & 54 / 57 \\
Negative entropy & 0.5 & 87.17 & 57.30 & 3 & 57 / 57 \\
\bottomrule
\end{tabular}

\end{table}

At these operating points, the learned logits probes have higher correct-abstention rates than MSP and the negative entropy.
The common framework can reuse different scorers, and the choice of regularization is a choice of operating point.
This table is a comparison of components: the threshold calibration and the executor of the $C=1$ probe are those of this paper, and the probe at its default probability threshold is listed in Table~\ref{tab:clip-logits}.

\subsection{A Common-Executor Control for CLIP Correlation Features}
\label{sec:exp-clip}

Appendix~F.2 of CLIP-UP gives a mechanism that abstains directly after a classification of the correlation vector~\citep{vardi2026clipup}.
We use the official Structure-CLIP weights~\citep{huang2023structureclip}, multiply the normalized image embedding element-wise with each of the four question--candidate text embeddings, and concatenate the products; the NOTA candidate is excluded.
The original paper does not give the complete fitting hyperparameters of this classifier, so we report two mechanism ports: an unstandardized fit with $C=1$, and a fit selected by cross-validation over image clusters within \setT.
The cross-validation chooses whether to standardize the features and a value of $C\in\{0.01, 0.1, 1, 10, 100\}$, and it selects the unstandardized fit with $C=100$.
The threshold rule and the executor are the same as for the logits probe (Table~\ref{tab:clip}).

\begin{table}[ht]
\centering
\caption{CLIP correlation-vector classifier on \Bone, on \setE, at its default threshold and under the calibration and executor of \method.
The calibrated rows are \method{} with the CLIP classifier as the scorer.}
\label{tab:clip}
\small
\begin{tabular}{@{}llrrrrl@{}}
\toprule
Classifier fit & Operating point & Standard (\%) & UPD (\%) & Lost & Retained & Joint criterion \\
\midrule
Fixed $C=1$ & Default threshold ($p=0.5$) & 48.99 & 92.40 & 2151 & 30 / 57 & not passed \\
 & Calibrated, $\varepsilon=0$ & 87.22 & 55.09 & 0 & 57 / 57 & passed \\
 & Calibrated, $\varepsilon=0.5\%$ & 86.70 & 57.46 & 29 & 55 / 57 & not passed \\
\midrule
Cross-validated & Default threshold ($p=0.5$) & 66.57 & 85.25 & 1162 & 51 / 57 & not passed \\
 & Calibrated, $\varepsilon=0$ & 87.22 & 54.75 & 0 & 57 / 57 & not passed \\
 & Calibrated, $\varepsilon=0.5\%$ & 87.08 & 56.59 & 8 & 57 / 57 & passed \\
\bottomrule
\end{tabular}

\end{table}

At the common 0.5\% budget, the cross-validated CLIP classifier and the cross-validated logits probe both lose exactly 8 Standard answers; the correct-abstention rate of the logits probe is 17.46 pp higher, with a 95\% interval of [15.71, 19.25].
The CLIP classifier retains all 57 SFT-gained records and the logits probe retains 55, which shows that the same total loss can correspond to different sets of lost records.
The classifier operating points that are not calibrated against the loss have high correct-abstention rates and veto a large number of original correct answers; after calibration, accuracy is retained, but the additional abstention gain is small.
At the low-loss operating points that we tested, this supports the internal logits score over the CLIP correlation features.

Table~\ref{tab:clip-logits} puts the default threshold next to the common calibration for the two cross-validated classifiers, all computed on the same \Bone.

\begin{table}[ht]
\centering
\caption{Default threshold against common calibration for the two cross-validated classifiers on \Bone, on \setE.}
\label{tab:clip-logits}
\small
\resizebox{\linewidth}{!}{\begin{tabular}{@{}llrrrrl@{}}
\toprule
Feature & Decision rule & Standard (\%) & UPD (\%) & Lost & Retained & Joint criterion \\
\midrule
CLIP correlation vector & Default threshold ($p=0.5$) & 66.57 & 85.25 & 1162 & 51 / 57 & not passed \\
 & Calibrated, $\varepsilon=0.5\%$ & 87.08 & 56.59 & 8 & 57 / 57 & passed \\
\midrule
First-token logits & Default threshold ($p=0.5$) & 81.92 & 92.16 & 298 & 45 / 57 & not passed \\
 & Calibrated, $\varepsilon=0.5\%$ & 87.08 & 74.05 & 8 & 55 / 57 & passed \\
\bottomrule
\end{tabular}
}
\end{table}

The default classification threshold does not retain the original accuracy with either feature; the common calibration against the loss brings both to operating points that pass the joint criterion.
Calibration also lowers abstention coverage, and the logits probe keeps more of the gain in correct-abstention rate at the calibrated operating point.

\subsection{A Multiple-Choice Adaptation of Adaptive Confidence Fusion}
\label{sec:exp-acr}

ACR fuses MSP with a correctness signal and a gate signal that are produced by two MLPs~\citep{tran2026acr}.
We adapt its inputs to the final-prompt hidden state of \Bone{} and the logits of the options A--E, keep the answer model frozen and the 15 epochs of training, and report one variant with raw features and one with features standardized using statistics of \setT{} only.
We call this adaptation ACR-MCQ.
Both variants use the same loss calibration and NOTA executor (Table~\ref{tab:acr}).

\begin{table}[ht]
\centering
\caption{ACR-MCQ on \Bone, on \setE.
The calibrated rows are \method{} with the ACR-MCQ score as the scorer.}
\label{tab:acr}
\small
\begin{tabular}{@{}llrrrr@{}}
\toprule
Features & Operating point & Standard (\%) & UPD (\%) & Lost & Retained \\
\midrule
Raw & Default threshold ($p=0.5$) & 85.76 & 67.00 & 82 & 25 / 57 \\
 & Calibrated, $\varepsilon=0$ & 87.22 & 55.56 & 0 & 57 / 57 \\
 & Calibrated, $\varepsilon=0.5\%$ & 86.97 & 59.32 & 14 & 54 / 57 \\
\midrule
Standardized & Default threshold ($p=0.5$) & 82.28 & 82.21 & 278 & 23 / 57 \\
 & Calibrated, $\varepsilon=0$ & 87.22 & 57.98 & 0 & 57 / 57 \\
 & Calibrated, $\varepsilon=0.5\%$ & 86.58 & 67.20 & 36 & 53 / 57 \\
\bottomrule
\end{tabular}

\end{table}

With raw inputs, the gate equals 1 on the whole training set, and the deployed behavior equals MSP over the option space A--E; standardization on \setT{} makes the gate vary with the input, and the correctness AUROC on \setE{} rises from 0.806 to 0.913.
At the 0.5\% calibration budget, the standardized variant has higher abstention coverage and also a larger evaluation-set loss.
Relative to this operating point, \methodrelaxed{} is 0.50 pp higher on Standard, with a 95\% interval of [0.21, 0.81], and 6.85 pp higher in correct-abstention rate, with an interval of [5.32, 8.32].

\subsection{A Visual-Contrast Correctness Probe}
\label{sec:exp-bicr}

BICR trains a shared MLP on the final-prompt states under the real image and under a black image, with the correctness of the actual answer of \Bone{} as the label and a joint objective of a weighted binary cross-entropy, a Brier term, and a ranking loss on positive examples~\citep{khanmohammadi2026bicr}.
Within \setT{} we make an 80/20 split by image cluster, run 50 trials of an Optuna search~\citep{akiba2019optuna}, and then train the selected configuration on the full \setT{} for 79 epochs.
\setCM{} is used only to calibrate the common loss threshold (Table~\ref{tab:bicr}).

\begin{table}[ht]
\centering
\caption{BICR on \Bone, on \setE.
The calibrated rows are \method{} with the BICR score as the scorer.}
\label{tab:bicr}
\small
\begin{tabular}{@{}lrrrrl@{}}
\toprule
Operating point & Standard (\%) & UPD (\%) & Lost & Retained & Joint criterion \\
\midrule
Default threshold ($p=0.5$) & 66.81 & 95.19 & 1148 & 22 / 57 & not passed \\
Calibrated, $\varepsilon=0$ & 87.22 & 55.48 & 0 & 57 / 57 & passed \\
Calibrated, $\varepsilon=0.5\%$ & 87.04 & 58.23 & 10 & 57 / 57 & passed \\
\bottomrule
\end{tabular}

\end{table}

The correctness AUROC of BICR on \setE{} is 0.904.
At the common 0.5\% budget it retains all SFT-gained records; the correct-abstention rate of \methodrelaxed{} is 15.82 pp higher, with a 95\% interval of [14.25, 17.41], and the difference on Standard is 0.04 pp in favor of \methodrelaxed, with an interval of [$-$0.14, 0.20].
The ranking of overall correctness and the abstention gain at low loss are different readouts.
BICR and ACR predict whether the answer is correct, whereas our probe and the CLIP classifier predict whether the input is answerable, with Standard against UPD as the classes; the differences among these complete scorers therefore combine features, targets, and fitting procedures.

\subsection{Budget Curves and Failure Types}
\label{sec:exp-curves}

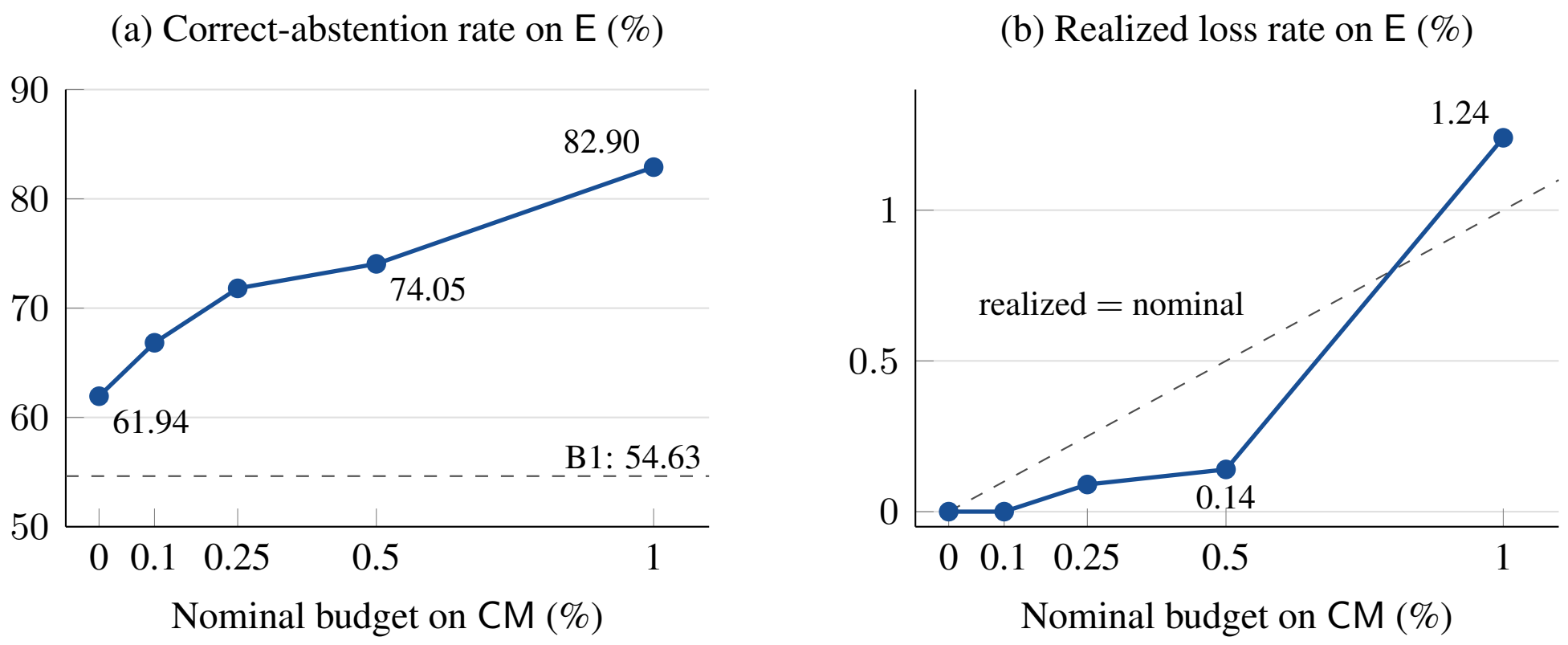
\begin{figure}[ht]
\centering
\definecolor{figink}{HTML}{184F95}
\begin{tikzpicture}
\begin{groupplot}[
  group style={group size=2 by 1, horizontal sep=1.7cm},
  scale only axis, width=5.3cm, height=3.6cm,
  xmin=-0.06, xmax=1.1, xtick={0,0.1,0.25,0.5,1}, xticklabels={0,0.1,0.25,0.5,1},
  xlabel={Nominal budget on \setCM{} (\%)},
  label style={font=\footnotesize}, ticklabel style={font=\footnotesize}, title style={font=\footnotesize},
  axis x line*=bottom, axis y line*=left,
  ymajorgrids, grid style={black!12},
  clip=false,
]
\nextgroupplot[title={(a) Correct-abstention rate on \setE{} (\%)}, ymin=50, ymax=90]
\draw[dashed, black!70] (axis cs:-0.06,54.63) -- (axis cs:1.1,54.63);
\node[font=\scriptsize, anchor=south east, inner sep=1.5pt] at (axis cs:1.1,54.63) {\Bone: 54.63};
\addplot[figink, line width=1pt, mark=*, mark size=1.8pt] coordinates {(0,61.94) (0.1,66.83) (0.25,71.81) (0.5,74.05) (1,82.90)};
\node[font=\scriptsize, anchor=north west, inner sep=3pt] at (axis cs:0,61.94) {61.94};
\node[font=\scriptsize, anchor=north west, inner sep=3pt] at (axis cs:0.5,74.05) {74.05};
\node[font=\scriptsize, anchor=south east, inner sep=3pt] at (axis cs:1,82.90) {82.90};
\nextgroupplot[title={(b) Realized loss rate on \setE{} (\%)}, ymin=-0.05, ymax=1.4, ytick={0,0.5,1}]
\draw[dashed, black!70] (axis cs:0,0) -- (axis cs:1.1,1.1);
\node[font=\scriptsize, anchor=south east, inner sep=2pt] at (axis cs:0.55,0.62) {realized $=$ nominal};
\addplot[figink, line width=1pt, mark=*, mark size=1.8pt] coordinates {(0,0.00) (0.1,0.00) (0.25,0.09) (0.5,0.14) (1,1.24)};
\node[font=\scriptsize, anchor=north, inner sep=3.5pt] at (axis cs:0.5,0.14) {0.14};
\node[font=\scriptsize, anchor=south east, inner sep=3pt] at (axis cs:1,1.24) {1.24};
\end{groupplot}
\end{tikzpicture}
\caption{Calibration budget against abstention gain and evaluation-set loss for the default scorer on \Bone.
The horizontal axis is the nominal budget on \setCM.
(a) Correct-abstention rate on \setE; the dashed line marks \Bone.
(b) Realized original-correct loss rate on \setE; on the dashed line the realized loss rate equals the nominal budget.
Empirical calibration does not force the evaluation points to lie below this line.}
\label{fig:budget}
\end{figure}

With the scorer fixed, we read the same threshold rule at \setCM{} budgets of 0\%, 0.1\%, 0.25\%, 0.5\%, and 1\%.
The correct-abstention rate of the default scorer on \setE{} is then 61.94\%, 66.83\%, 71.81\%, 74.05\%, and 82.90\%, and the number of original correct answers lost is 0, 0, 5, 8, and 70.
A wider budget increases abstention coverage and also the realized loss.
The curve is the single-parameter frontier of Property~\ref{prop:onesided}(c); it shows the trade-off among operating points, and the preset primary and additional operating points stay unchanged (Figure~\ref{fig:budget}).
At the 1\% budget, the realized loss rate on \setE, 1.24\%, is above the nominal budget.

Table~\ref{tab:tasks} separates the comparison by failure type.
Relative to EmbLoRA, the advantage of \methodrelaxed{} in correct-abstention rate is concentrated in AAD and IASD.
On IVQD, EmbLoRA has a higher correct-abstention rate and corrects 9 errors of \Bone, so that its Standard accuracy on this subtask is also higher; \methodrelaxed{} retains all 13 SFT-gained records and loses 1 original correct answer.
Different failure types give different trade-offs among answering, abstention, and retention.

\begin{table}[ht]
\centering
\caption{Results by failure type on \Bone, on \setE.}
\label{tab:tasks}
\small
\begin{tabular}{@{}llrrrr@{}}
\toprule
Task & Scheme & Standard (\%) & UPD (\%) & Lost & Retained \\
\midrule
AAD & \methodrelaxed & 88.32 & 60.47 & 3 & 20 / 21 \\
 & CLIP-UP EmbLoRA & 88.01 & 38.80 & 25 & 14 / 21 \\
 & Plain abstention SFT & 86.39 & 80.39 & 58 & 7 / 21 \\
\midrule
IASD & \methodrelaxed & 86.31 & 85.38 & 4 & 22 / 23 \\
 & CLIP-UP EmbLoRA & 85.87 & 77.35 & 32 & 15 / 23 \\
 & Plain abstention SFT & 84.24 & 94.09 & 70 & 8 / 23 \\
\midrule
IVQD & \methodrelaxed & 86.14 & 68.36 & 1 & 13 / 13 \\
 & CLIP-UP EmbLoRA & 86.67 & 76.83 & 5 & 11 / 13 \\
 & Plain abstention SFT & 85.61 & 82.65 & 12 & 8 / 13 \\
\bottomrule
\end{tabular}

\end{table}

\subsection{Frozen Confirmation on Independent Images}
\label{sec:exp-confirm}

On the confirmation set, the Standard accuracy of \Bone{} is 91.67\% and its correct-abstention rate is 21.69\%.
Both frozen operating points of \method{} keep all 836 original correct answers of \Bone, an original-correct loss of 0 of 912; the correct-abstention rate rises by 20.10 pp, with a 95\% interval of [16.11, 24.64], and by 44.36 pp, with an interval of [39.30, 50.00], respectively.
Both operating points pass the joint criterion (Table~\ref{tab:confirm}).

\begin{table}[ht]
\centering
\caption{Frozen confirmation on \setI.
UPD is given as the micro-average over the 816 UPD records and as the macro-average over the failure types.
The joint criterion is relative to \Bone; lost and corrected answers are counted among the 912 Standard records, and retained records are out of the 8 SFT-gained records.}
\label{tab:confirm}
\small
\resizebox{\linewidth}{!}{\begin{tabular}{@{}lrrrrrl@{}}
\toprule
Branch & Standard (\%) & UPD, micro (\%) & UPD, macro (\%) & Lost / corrected & Retained & Joint criterion \\
\midrule
\Bone & 91.67 & 21.69 & 21.29 & 0 / 0 & 8 / 8 & reference \\
\method & 91.67 & 41.79 & 37.95 & 0 / 0 & 8 / 8 & passed \\
\methodrelaxed & 91.67 & 66.05 & 60.10 & 0 / 0 & 8 / 8 & passed \\
CLIP-UP EmbLoRA & 92.32 & 33.33 & 33.33 & 7 / 13 & 6 / 8 & passed \\
Plain abstention SFT & 91.78 & 70.96 & 66.75 & 4 / 5 & 6 / 8 & passed \\
\bottomrule
\end{tabular}
}
\end{table}

Relative to EmbLoRA, the correct-abstention rate of \method{} is 8.46 pp higher, with an interval of [3.58, 13.73], and that of \methodrelaxed{} is 32.72 pp higher, with an interval of [27.13, 38.53].
Both operating points lose 7 fewer original correct answers, a reduction of 0.77 pp in the loss rate, with an interval of [0.22, 1.43].
EmbLoRA also corrects 13 old errors, which puts its net Standard accuracy 0.66 pp above that of \method; the interval of the Standard difference, \method{} minus EmbLoRA, is [$-$1.87, 0.44].

Plain abstention SFT also passes the joint criterion in this round.
Relative to \methodrelaxed, its correct-abstention rate is 4.90 pp higher, with an interval of [1.34, 8.49], and it loses 4 original correct answers and corrects 5 old errors.
The original-correct loss rate of \method{} is 0.44 pp lower, with an interval of [0.11, 0.88]; the net difference on Standard is $-$0.11 pp, with an interval of [$-$0.88, 0.55].
The independent confirmation thus continues the trade-off between retention and abstention coverage: \method{} keeps all original correct answers, plain abstention SFT has higher abstention coverage, and EmbLoRA brings more additional corrections.

By task, the correct-abstention rates of \methodrelaxed{} on AAD and IASD are 60.07\% and 86.52\%, above the 8.68\% and 48.88\% of EmbLoRA.
On IVQD, \methodrelaxed{} is at 33.72\%, and EmbLoRA and plain abstention SFT are at 42.44\% and 49.42\%, in agreement with the relative trade-off on \setE.
Relative to \Bzero, \Bone{} gains 8 records and loses 4 on the confirmation set; the SFT-gained records lie in 7 clusters, \method{} retains all of them at both operating points, and each of the two trained branches retains 6.
Only counts are reported for this small set.
The complete joint criterion, the four direct comparisons, Circular Dual, and the per-task values are given in Appendix~\ref{app:confirm}.

\section{Discussion}
\label{sec:discussion}

\paragraph{The retention requirement of abstention.}
When net accuracies are close, abstention branches can still change different numbers of original correct answers.
The difference between \methodrelaxed{} and EmbLoRA on \Bone{} shows that the decomposition into losses and corrections complements the net metric.
For applications that continue to rely on existing answers, a deployer can explicitly choose a stricter retention point, or accept a small loss for higher abstention coverage.

\paragraph{The frozen setting on new images.}
On \setE{} and on the independent confirmation set \setI, both preset operating points of \method{} give a low original-correct loss and additional abstention.
Plain abstention SFT loses 140 of 5,626 on \setE{} and 4 of 912 on \setI, and passes the joint criterion on the latter; the retention cost of the trained branches varies with the evaluation population.
On \setI, EmbLoRA reaches a higher net Standard accuracy through corrections, while \method{} obtains more correct abstentions and keeps all original correct answers, which again shows the difference between net accuracy and the retention of behavior.

\paragraph{Post-training changes the abstention operating point.}
The same probe pipeline has a lower correct-abstention rate on \Bone{} than on \Bzero: 67.85\% becomes 61.94\% at the zero budget, and 80.56\% becomes 74.05\% at the 0.5\% budget.
The abstention layer therefore needs to be evaluated again after answer adaptation.
The drop in the operating point alone does not yet separate the effects of a change in ranking, a change in calibration, and the reuse of the training set; the score distributions on \setT{} and \setCM{} and the ranking at low false-positive rates could separate these explanations further.

\paragraph{Choice of compared methods.}
CLIP-UP provides a route of training and injection that is aimed directly at UPD; LVLM-LP is the direct source of the internal logits score; VL-Uncertainty provides a perturbation-based uncertainty control; MSP and the negative entropy are simple scorers that need no additional supervised learning.
The correlation-vector classifier in Appendix~F.2 of CLIP-UP is the closest component control; BICR adds a visual-contrast correctness probe, and ACR-MCQ adds a learnable confidence fusion.
The comparison is organized by task, mechanism, and reproducibility, and does not merge numbers that papers report on different datasets into a single ranking.

\section{Limitations}
\label{sec:limitations}

The shared-SFT setting has a single training seed, and the training set has 677 records.
The net Standard gain of SFT is a point estimate whose interval contains zero; the 57 SFT-gained records lie in 28 image clusters, and the interval of the difference in gain retention rate is wide.
The image-cluster intervals of this paper measure sampling variation under fixed checkpoints and do not include the randomness of retraining.

\setE{} is a historically exposed evaluation population, and \setI{} provides a confirmation on independent images after the method was frozen.
\setI{} covers only AI2D science diagrams, the three failure types have different sample sizes, and the questions and transformations were reviewed by rules and by a multimodal model, without annotation by two human annotators.
That the images were not used in the development of this project does not exclude that the base model saw public AI2D data during pretraining or instruction tuning.
The abstention baseline of \Bone{} is lower on \setI{} than on \setE, so the absolute gains on the two populations cannot be read directly as a change in the ability of the method; we interpret the results through the paired comparisons within each set.
On \setI{} there are only 8 SFT-gained records in 7 clusters, and their retention counts do not support a new statistical conclusion.

The preceding SFT and the probe share the Standard records of \setT.
The budget on \setCM{} is an empirical constraint.
Property~\ref{prop:newcluster} covers only the zero budget and a single new image cluster that is exchangeable with the calibration clusters; the zero loss of the two operating points on the confirmation set is not covered by it and gives no population-level guarantee.
For \setE, the exchangeability of its image clusters with those of \setCM{} is an assumption about how the data were partitioned.
The results on the two original base models support fitting the pipeline separately on each base model; the results on retention after post-training currently come only from Qwen2.5-VL-3B.

In an experiment with variations of the frozen prompt, part of the gain shrinks; these runs are not reported here.
The direct retention comparison for abstention by parameter update covers EmbLoRA and plain abstention SFT; the two differ in target text and prompt format, so plain abstention SFT is not a causal ablation that removes only the CLIP signal.
The CLIP classifier result is a mechanism port under the common prompt, fitted over a disclosed finite grid of candidates.
BICR keeps its core objective and search scheme and is adapted to the common \Bone, data, and executor.
ACR-MCQ changes the inputs and the head width of the original audio-visual classification model to the setting of this task, and the standardized adaptation is reported in separate rows.
The added controls were run after the main results had been observed and are extension experiments.
Rep2Act~\citep{rep2act2026}, Reliable-V-RL~\citep{jian2026reliablevrl}, VL-Calibration~\citep{xiao2026vlcalibration}, Dr.~No~\citep{hasanov2026drno}, and the additional-instruction prompting of MM-UPD~\citep{miyai2025upd} have no results in the same setting; Appendix~\ref{app:fidelity} gives the state of each.

\section{Conclusion}
\label{sec:conclusion}

On the present multimodal abstention task, a constraint on the loss of correct answers, combined with deterministic execution, turns an existing answerability score into favorable operating points between abstention and retention.
Record-level evaluation from a shared answer-SFT starting point shows that the abstention stage can keep all or most of the SFT-gained answers while adding correct abstentions.
The confirmation on independent images, run after the models and thresholds were frozen, further shows that the two preset operating points raise the correct-abstention rate while keeping all original correct answers.
Losses, corrections, and gain retention, together with net accuracy, show the actual trade-offs of the different abstention routes.

\bibliographystyle{plainnat}
\bibliography{refs}

\appendix
\section{Proofs}
\label{app:proofs}

Properties~\ref{prop:onesided} and~\ref{prop:loss} follow directly from the veto rule of Eq.~\eqref{eq:veto}, as explained after their statements in Section~\ref{sec:properties}.
This appendix proves Property~\ref{prop:newcluster} and adds three remarks on its scope.

\paragraph{Setup.}
An image cluster is a finite set of records, and the score $s$ is real-valued.
For an image cluster $k$, $m_k\in(-\infty,+\infty]$ is the lowest score among its Standard records that $f$ answers correctly, with $m_k=+\infty$ if there is no such record.
Clusters $1,\dots,K$ are the image clusters of \setCM, cluster $K+1$ is the new image cluster, and $M_K=\min_{k\le K}m_k$.

\paragraph{The zero-budget threshold.}
By Property~\ref{prop:loss}, $g_\tau$ loses a Standard record of \setCM{} that $f$ answers correctly exactly when the score of the record is below $\tau$.
Hence $\widehat{L}_{\setCM}(g_\tau;f)=0$ if and only if $\tau\le M_K$, and the rule of Eq.~\eqref{eq:calibrate} gives
\begin{equation}
\tau_0\le M_K .
\label{eq:tau0}
\end{equation}
The candidate set includes $-\infty$, which is always feasible and vetoes no record.
The proofs below use only Eq.~\eqref{eq:tau0} and therefore hold for any candidate set $\cand$ with an always-feasible no-veto endpoint.
In the same way, $g_{\tau_0}$ loses an original correct answer in the new cluster exactly when $m_{K+1}<\tau_0$, which implies $m_{K+1}<M_K$.

\begin{proof}[Proof of part (a)]
For $i=1,\dots,K+1$, let $A_i$ be the event that $m_i<m_j$ for all $j\ne i$.
The events $A_i$ are pairwise disjoint, so $\sum_{i}\Pr[A_i]\le 1$.
Under the conditioning of the property, $f$ and $s$ are fixed, each $m_k$ is the same function of cluster $k$, and the clusters are exchangeable; the values $m_1,\dots,m_{K+1}$ are therefore exchangeable, and all $\Pr[A_i]$ are equal.
Hence $\Pr[A_{K+1}]\le 1/(K+1)$.
A loss in the new cluster implies $m_{K+1}<M_K$, which is the event $A_{K+1}$.
If the lowest value is attained by more than one cluster, including the case in which several clusters have $m_k=+\infty$, no $A_i$ occurs, which only lowers the probability of a loss.
The bound holds under the conditioning and therefore also after averaging over it.
\end{proof}

\begin{proof}[Proof of part (b)]
Under the conditioning, $m_1,\dots,m_{K+1}$ are i.i.d.
For $t\in[-\infty,+\infty]$, let $G(t)=\Pr[m_1<t]$.
$G$ is non-decreasing and left-continuous, with $G(-\infty)=0$ and $G(+\infty)=\Pr[m_1<+\infty]$.
Because the clusters are i.i.d., conditioning on the calibration clusters does not change the distribution of the new cluster; the conditional probability of a loss in the new cluster is therefore $G(\tau_0)$, and $G(\tau_0)\le G(M_K)$ by Eq.~\eqref{eq:tau0}.
It remains to show that $\Pr[G(M_K)>\alpha]\le(1-\alpha)^K$.
If $G(+\infty)\le\alpha$, then $G(M_K)\le\alpha$ always.
Otherwise, let $t^{\ast}=\sup\{t\in\mathbb{R}: G(t)\le\alpha\}$.
This set is non-empty because $G(t)\to 0$ as $t\to-\infty$, and it is bounded above because $G(t)\to G(+\infty)>\alpha$ as $t\to+\infty$; hence $t^{\ast}$ is finite.
By left-continuity, $G(t^{\ast})\le\alpha$.
Since $G(t)>\alpha$ for every $t>t^{\ast}$, $\Pr[m_1\le t^{\ast}]=\lim_{t\downarrow t^{\ast}}G(t)\ge\alpha$.
If $G(M_K)>\alpha$, then $M_K>t^{\ast}$, that is, $m_k>t^{\ast}$ for all $k\le K$; this event has probability $(1-\Pr[m_1\le t^{\ast}])^K\le(1-\alpha)^K$.
\end{proof}

For the level quoted in Section~\ref{sec:properties}, setting $(1-\alpha)^{K}=0.05$ with $K=183$ gives $\alpha=1-0.05^{1/183}\approx 1.62\%$.

\paragraph{Remark 1: several evaluation clusters.}
Consider an evaluation set with $N$ image clusters, each of which is exchangeable with the $K$ calibration clusters in the sense of the property.
By part~(a) and the linearity of expectation, the expected number of evaluation clusters in which an original correct answer is lost is at most $N/(K+1)$; for \setE{} this is $416/184\approx 2.26$.
The loss events of different evaluation clusters share the threshold $\tau_0$; the marginal bound alone does not establish concentration of this count around its expectation.
The probability that at least one of the $N$ clusters has a loss is not small.
If all $N+K$ clusters are exchangeable, $\tau_0=M_K$, and the values $m_k$ are almost surely distinct, a loss occurs in some evaluation cluster exactly when the lowest of the $N+K$ values belongs to an evaluation cluster, which has probability $N/(N+K)=416/599\approx 69.4\%$.

\paragraph{Remark 2: budgets above zero.}
The same argument gives a statement at the level of clusters for a budget of $r$ calibration records.
If $g_\tau$ loses at most $r$ records of \setCM, then at most $r$ calibration clusters have $m_k<\tau$, so the threshold does not exceed the $(r+1)$-th smallest of $m_1,\dots,m_K$.
A loss in the new cluster then requires that at most $r$ of the other $K$ values be less than or equal to $m_{K+1}$.
At most $r+1$ of the $K+1$ clusters can have this property at the same time, because the cluster with the largest value among them has all the others at or below it; by exchangeability, the probability is at most $(r+1)/(K+1)$.
For $\varepsilon=0.5\%$, $r=12$, and the bound is $13/184\approx 7.07\%$.
This is the probability that a new cluster contains at least one lost record; it is loose, and it is not a bound on the loss rate over records.
We therefore state Property~\ref{prop:newcluster} for the zero budget only and base no claim about \methodrelaxed{} on this remark.

\paragraph{Remark 3: what the property does not cover.}
The bound is marginal over image clusters and is not conditional on a failure type or on any other subgroup.
It concerns the loss only and is silent on utility: a rule that never vetoes satisfies it, and the gain in correct-abstention rate at $\tau_0$ is an empirical matter.
It is tied to the frozen $f$, the frozen scorer, the prompt, the option format, and the NOTA mapping $n(\cdot)$; if any of these changes after calibration, the threshold has to be calibrated again.

\section{Calibration Budget and Evaluation-Set Loss}
\label{app:budget}

On \Bone, the zero-budget operating point of every scorer has an original-correct loss of 0 on both \setCM{} and \setE.
Table~\ref{tab:budget} gives the operating points at the 0.5\% budget; the denominators are the 2,406 Standard records of \setCM{} and the 5,626 Standard records of \setE.

\begin{table}[ht]
\centering
\caption{Original-correct loss at the 0.5\% calibration budget on \Bone, on the calibration set and on the evaluation set.
Every row is \method{} with the named scorer.}
\label{tab:budget}
\small
\begin{tabular}{@{}lrrrr@{}}
\toprule
Scorer & Losses on \setCM & Loss rate on \setCM{} (\%) & Losses on \setE & Loss rate on \setE{} (\%) \\
\midrule
Cross-validated probe (default) & 11 & 0.46 & 8 & 0.14 \\
Probe with fixed $C=1$ & 12 & 0.50 & 13 & 0.23 \\
MSP & 7 & 0.29 & 14 & 0.25 \\
Negative entropy & 8 & 0.33 & 3 & 0.05 \\
CLIP classifier, fixed $C=1$ & 12 & 0.50 & 29 & 0.52 \\
CLIP classifier, cross-validated & 12 & 0.50 & 8 & 0.14 \\
ACR-MCQ, raw features & 7 & 0.29 & 14 & 0.25 \\
ACR-MCQ, standardized features & 10 & 0.42 & 36 & 0.64 \\
BICR & 11 & 0.46 & 10 & 0.18 \\
\bottomrule
\end{tabular}

\end{table}

The loss rates on \setE{} of the CLIP classifier with fixed $C=1$ and of the standardized ACR-MCQ are above the nominal calibration budget.
The empirical constraint on the calibration set and the realized value on the evaluation set therefore have to be reported separately.
Table~\ref{tab:budget-curve} gives the complete counts for the five budgets of Section~\ref{sec:exp-curves}.

\begin{table}[ht]
\centering
\caption{The default scorer on \Bone{} at five calibration budgets, on \setE.}
\label{tab:budget-curve}
\small
\begin{tabular}{@{}rrrr@{}}
\toprule
Nominal budget on \setCM{} (\%) & Correct-abstention rate on \setE{} (\%) & Losses on \setE & Loss rate on \setE{} (\%) \\
\midrule
0 & 61.94 & 0 & 0.00 \\
0.1 & 66.83 & 0 & 0.00 \\
0.25 & 71.81 & 5 & 0.09 \\
0.5 & 74.05 & 8 & 0.14 \\
1 & 82.90 & 70 & 1.24 \\
\bottomrule
\end{tabular}

\end{table}

\section{Implementation and Fidelity of the Compared Methods}
\label{app:fidelity}

This appendix gives, for each compared method, its fidelity label in the sense of Section~\ref{sec:protocol} and the points at which our implementation differs from the original.
All methods use the data roles of Table~\ref{tab:protocol}, the common option prompt, and the common parser.

\paragraph{LVLM-LP probe.}
Mechanism port.
The feature, namely the full-vocabulary logits at the first answer position, and the linear form of the probe follow LVLM-LP and its official repository~\citep{zhao2024lvlmlp}.
The regularization strength is selected by cross-validation over image clusters within \setT, and the fit with fixed $C=1$ is reported as a second scorer.
Table~\ref{tab:clip-logits} lists the probe at its default probability threshold; all calibrated rows are \method.

\paragraph{CLIP-UP.}
On InternVL3-8B, a base model supported by the original work, we use the released code and the training recipe of the authors~\citep{vardi2026clipup}, train on \setT, and evaluate under the common option prompt.
On Qwen2.5-VL-3B, for both \Bzero{} and \Bone, the implementation is a mechanism port, because the released code does not cover this model family.
The projected correlation token is placed after the vision-end token, whereas the official InternVL implementation places it before the image-end token.
Training, fitting, and calibration are done separately on each base model.
The Emb result on InternVL3-8B in Table~\ref{tab:base} is a result of transferring the method to the present protocol; it is not a verdict on the numbers of the original paper, which were obtained with its own data and prompts.

\paragraph{VL-Uncertainty.}
Mechanism port of the official repository~\citep{zhang2024vluncertainty}, with the model interface changed from Qwen2-VL to Qwen2.5-VL; we report it on the original Qwen2.5-VL-3B only.
Each record is answered under five perturbed versions of the input, which pair a blurred image with a rephrased question, and the uncertainty is the base-2 entropy of the answer clusters over the five samples.
In the native execution, a record with an uncertainty of at least 1.0 receives the displayed abstention letter, and every other record receives the initial answer, which is sampled at temperature 0.1.
Under the common executor, the uncertainty is used as the score, and the cached greedy answer of the base model is kept on the answered side.
Three deviations from the official code are disclosed.
The full question and hint are rephrased, where the official code rephrases the first line only.
Answers are clustered by the frozen option parser of the protocol, where the official code takes the first integer.
Records whose questions have an identical prefix share their rephrasings.

\paragraph{MSP and negative entropy.}
Both are computed from the output distribution of $f$ and need no fitting; they have no threshold of their own and appear only with the calibration and executor of \method.

\paragraph{CLIP correlation-vector classifier.}
Mechanism port of the classifier in Appendix~F.2 of CLIP-UP, with the official Structure-CLIP weights~\citep{huang2023structureclip}.
The original paper does not give the complete fitting hyperparameters, so two fits are disclosed (Section~\ref{sec:exp-clip}).

\paragraph{BICR.}
Mechanism port.
The core objective and the search scheme of BICR are kept~\citep{khanmohammadi2026bicr}, and the method is adapted to \Bone, to \setT, and to the common executor.

\paragraph{ACR-MCQ.}
An adaptation of ACR, which was proposed for an audio-visual classification model~\citep{tran2026acr}.
The inputs and the width of the heads are changed to the setting of this task (Section~\ref{sec:exp-acr}), and we therefore report it under a separate name.

\paragraph{Plain abstention SFT.}
This branch is defined in this paper (Section~\ref{sec:exp-b1}) and has no original implementation.

\paragraph{Methods without results in the same setting.}
Rep2Act trains a routing head together with rationale targets that are grounded in the image~\citep{rep2act2026}.
We have implemented the head, the four training objectives, and the two-stage training; \setT{} has no such rationales, so a run on it would be a partial implementation, and none is reported.
Reliable-V-RL trains with rewards from a grader model that scores responses against rubrics written for each question~\citep{jian2026reliablevrl}.
We have implemented the reward and the training; the rubrics, the contradiction statements from which they are written, and the grader model are not available for \setT, and no run is reported.
For VL-Calibration~\citep{xiao2026vlcalibration}, the official source has been adapted and its data checks pass, but training has not been run.
Dr.~No needs a sparse autoencoder that matches the answer model~\citep{hasanov2026drno}; there is none for \Bone.
The additional-instruction prompting of MM-UPD~\citep{miyai2025upd} has not been run on \Bone.

\section{Loss of SFT-Gained and of Persistently Correct Records}
\label{app:repairs}

On \setE, \Bone{} has 57 SFT-gained records and a further 4,850 Standard records that both \Bzero{} and \Bone{} answer correctly, which we call persistently correct.
Table~\ref{tab:repairs} compares the rates at which the abstention stage loses the two kinds of records; the intervals use 2,000 paired bootstrap resamples over image clusters.

\begin{table}[ht]
\centering
\caption{Loss of SFT-gained records and of persistently correct records in the abstention stage, on \setE.}
\label{tab:repairs}
\small
\resizebox{\linewidth}{!}{\begin{tabular}{@{}lrrr@{}}
\toprule
Scheme & SFT-gained records lost (\%) & Persistently correct records lost (\%) & Difference [95\% interval] (pp) \\
\midrule
\methodrelaxed & 2 / 57 (3.51) & 6 / 4,850 (0.12) & 3.39 [$-$0.22, 12.25] \\
CLIP-UP EmbLoRA & 17 / 57 (29.82) & 45 / 4,850 (0.93) & 28.90 [12.82, 49.88] \\
Plain abstention SFT & 34 / 57 (59.65) & 106 / 4,850 (2.19) & 57.46 [37.68, 79.47] \\
\bottomrule
\end{tabular}
}
\end{table}

EmbLoRA and plain abstention SFT lose SFT-gained records at a higher rate, with intervals above zero; the interval of the difference for \methodrelaxed{} contains zero.
The set of SFT-gained records is itself selected by correctness before and after SFT and differs in difficulty from the persistently correct set; these statistics describe a conditional difference in behavior on the present sample.

\section{Construction of the Independent Confirmation and Full Comparisons}
\label{app:confirm}

\paragraph{Images and duplicates.}
Among the original AI2D questions, we use only those that need no added letter overlay on the image, and we keep the official answers and the original images.
Candidate images are matched, for exact and near duplicates, against the images used earlier in this project, all official AI2D test images, and the images of Geometry3K; the near-duplicate criterion is a Hamming distance of at most 4 between 64-bit dHash values.
Different confirmation clusters share neither original images nor replacement images.

\paragraph{Semantic review.}
The review of semantics uses fixed rules and a multimodal model.
Its inputs are limited to the original question, the options, the image, and the transformation; they do not include the answers, scores, or correctness of the branches under evaluation.
The 300 candidate clusters that entered the review were screened as in Table~\ref{tab:screening}, and the original correct answers were not modified.
All decisions were recorded, and the final sample was fixed before inference.

\begin{table}[ht]
\centering
\caption{Semantic screening of the candidate clusters of \setI.
Numbers in parentheses are counts of excluded clusters.}
\label{tab:screening}
\small
\begin{tabular}{@{}lrrrL{8.4cm}@{}}
\toprule
Type & Reviewed & Excluded & Final & Reasons for exclusion \\
\midrule
AAD & 100 & 4 & 96 & Options overlap in meaning, a meta-option cannot be judged, the stem reveals the answer, or the original question is unclear \\
IASD & 100 & 11 & 89 & Nested option labels (3); an ``NA'' meta-option (3); still answerable or ambiguous after replacement (3); a synonymous correct answer (1); confused concepts in the original question (1) \\
IVQD & 100 & 57 & 43 & Insufficient dependence on the image or an ambiguous question (55); the original answer is not unique on review of the image (2) \\
\bottomrule
\end{tabular}

\end{table}

\paragraph{Records and resampling unit.}
Every Standard question keeps four substantive options and is rotated over all circular shifts; AAD rotates the three options that remain after the correct one is removed, and IASD and IVQD rotate four options.
The NOTA text is ``None of the above'' for AAD and IASD and ``The image and question are irrelevant.'' for IVQD, and it is fixed in the last position.
The unit of resampling consists of all Standard and UPD records of an original question together with its unique foreign image, which gives 228 clusters.
Inference uses a batch size of 1, greedy decoding, BF16 precision, and at most 1,024 new tokens.

\paragraph{Full comparisons.}
The models and thresholds are those of the frozen configuration, and \setI{} is not used to select methods or operating points.
Table~\ref{tab:confirm-joint} gives the joint criterion with its intervals, and Table~\ref{tab:confirm-contrasts} reports all four direct comparisons that were fixed before inference.
Table~\ref{tab:confirm-tasks} gives the results by failure type.

\begin{table}[ht]
\centering
\caption{Joint criterion on \setI, relative to \Bone.
The lost count is out of 912 Standard records; ``Invalid'' counts outputs that the parser cannot map to an option.}
\label{tab:confirm-joint}
\small
\resizebox{\linewidth}{!}{\begin{tabular}{@{}lrrrrr@{}}
\toprule
Branch & $\Delta$Standard [95\% interval] (pp) & $\Delta$UPD [95\% interval] (pp) & Lost (\%) & Circular Dual (\%) & Invalid \\
\midrule
\Bone & reference & reference & 0 (0.00) & 10.08 & 0 \\
\method & 0.00 [0.00, 0.00] & 20.10 [16.11, 24.64] & 0 (0.00) & 23.93 & --- \\
\methodrelaxed & 0.00 [0.00, 0.00] & 44.36 [39.30, 50.00] & 0 (0.00) & 44.94 & --- \\
CLIP-UP EmbLoRA & 0.66 [$-$0.44, 1.87] & 11.64 [8.30, 15.17] & 7 (0.77) & 20.05 & 1 \\
Plain abstention SFT & 0.11 [$-$0.55, 0.88] & 49.26 [44.00, 54.63] & 4 (0.44) & 52.82 & 0 \\
\bottomrule
\end{tabular}
}
\end{table}

\begin{table}[ht]
\centering
\caption{The four direct comparisons on \setI{} that were fixed before inference, as differences with paired 95\% intervals over image clusters.
A positive $\Delta$Standard or $\Delta$UPD means that \method{} is higher; a positive reduction in original-correct loss means that \method{} loses fewer answers.}
\label{tab:confirm-contrasts}
\small
\resizebox{\linewidth}{!}{\begin{tabular}{@{}lrrr@{}}
\toprule
Comparison & $\Delta$Standard (pp) & $\Delta$UPD (pp) & Reduction in original-correct loss (pp) \\
\midrule
\method{} $-$ CLIP-UP EmbLoRA & $-$0.66 [$-$1.87, 0.44] & 8.46 [3.58, 13.73] & 0.77 [0.22, 1.43] \\
\method{} $-$ plain abstention SFT & $-$0.11 [$-$0.88, 0.55] & $-$29.17 [$-$34.10, $-$24.26] & 0.44 [0.11, 0.88] \\
\methodrelaxed{} $-$ CLIP-UP EmbLoRA & $-$0.66 [$-$1.87, 0.44] & 32.72 [27.13, 38.53] & 0.77 [0.22, 1.43] \\
\methodrelaxed{} $-$ plain abstention SFT & $-$0.11 [$-$0.88, 0.55] & $-$4.90 [$-$8.49, $-$1.34] & 0.44 [0.11, 0.88] \\
\bottomrule
\end{tabular}
}
\end{table}

\begin{table}[ht]
\centering
\caption{Results by failure type on \setI.
The numbers of Standard and UPD records are 384 and 288 for AAD, 356 and 356 for IASD, and 172 and 172 for IVQD.}
\label{tab:confirm-tasks}
\small
\begin{tabular}{@{}llrrr@{}}
\toprule
Task & Branch & Standard (\%) & UPD (\%) & Lost / corrected \\
\midrule
AAD & \Bone & 92.19 & 11.11 & 0 / 0 \\
 & \method & 92.19 & 28.47 & 0 / 0 \\
 & \methodrelaxed & 92.19 & 60.07 & 0 / 0 \\
 & CLIP-UP EmbLoRA & 94.01 & 8.68 & 1 / 8 \\
 & Plain abstention SFT & 92.97 & 63.19 & 1 / 4 \\
\midrule
IASD & \Bone & 95.79 & 29.49 & 0 / 0 \\
 & \method & 95.79 & 60.96 & 0 / 0 \\
 & \methodrelaxed & 95.79 & 86.52 & 0 / 0 \\
 & CLIP-UP EmbLoRA & 95.22 & 48.88 & 4 / 2 \\
 & Plain abstention SFT & 95.22 & 87.64 & 2 / 0 \\
\midrule
IVQD & \Bone & 81.98 & 23.26 & 0 / 0 \\
 & \method & 81.98 & 24.42 & 0 / 0 \\
 & \methodrelaxed & 81.98 & 33.72 & 0 / 0 \\
 & CLIP-UP EmbLoRA & 82.56 & 42.44 & 2 / 3 \\
 & Plain abstention SFT & 81.98 & 49.42 & 1 / 1 \\
\bottomrule
\end{tabular}

\end{table}

On \setI, the Standard accuracy of \Bzero{} is 91.23\% and that of \Bone{} is 91.67\%; SFT gains 8 records and loses 4.
The SFT-gained records lie in 7 clusters, below the threshold of 10 clusters that was set in advance for reporting an interval.

\end{document}